\documentclass[preprint,3p,times,twocolumn,authoryear]{elsarticle}

\usepackage{amssymb}
\usepackage{amsmath}
\usepackage{lineno}

\usepackage{lipsum}
\usepackage[dvipsnames,svgnames,x11names]{xcolor}
\usepackage{adjustbox}
\usepackage{booktabs}
\usepackage{hyperref}
\usepackage{orcidlink}
\usepackage{booktabs}
\usepackage[table,xcdraw]{xcolor}
\usepackage{svg}

\usepackage[markup=underlined]{changes}
\definechangesauthor[name={Jiapan}, color=orange]{JW}
\definechangesauthor[name={A}, color=BrickRed]{A}
\definechangesauthor[name={B}, color=NavyBlue]{B}
\definechangesauthor[name={Daan}, color=purple]{DH}

\usepackage{todonotes}
\setcommentmarkup{\todo[color={authorcolor!20},size=\scriptsize]{#3: #1}}

\journal{ISPRS Journal}

\begin{document}

\begin{frontmatter}



%% Article title
\title{DeepTopoClustering: Unsupervised Derivation of Surface Process Taxonomy from 4D Point Clouds for Topographic Monitoring}


\author[RSA]{Jiapan Wang} %%\orcidlink{0009-0003-0487-9729}
\author[TUD]{Daan Hulskemper}
\author[RSA]{Mathilde Letard}
\author[TUD]{Roderik Lindenbergh}
\author[RSA]{Katharina Anders}

\affiliation[RSA]{organization={Remote Sensing Applications, TUM School of Engineering and Design, Technical University of Munich},%Department and Organization
            city={Munich},
            country={Germany}}

\affiliation[TUD]{organization={Department of Geoscience \& Remote Sensing, Delft University of Technology},%Department and Organization
            city={Delft},
            country={The Netherlands}}       

% \affiliation[xx]{organization={xxx},
%             country={Germany}}

%% Abstract
\begin{abstract}
4D point clouds acquired by permanent laser scanning (PLS) enable accurate high-frequency monitoring of surface change in dynamic topographic environments. However, the resulting data volumes exceed the interpretive capacity of existent methods for identifying the underlying surface processes. Although these methods can detect and quantify where and when surface change occurs, they remain limited in organizing detected activities into meaningful process types. A sound methodology for identifying, comparing, and describing a variety of surface processes is therefore needed to support long-term topographic monitoring. To address this gap, we propose DeepTopoClustering (DTC), an unsupervised framework for deriving a hierarchical process taxonomy from object-based surface activities, so-called 4D objects-by-change (4D-OBCs). We first transform each 4D-OBC into a GeoMorphogram, a distributional sequence that represents the temporal evolution of topographic change within a spatially bounded surface activity. A convolutional autoencoder then learns latent embeddings from GeoMorphograms, which are then jointly optimized using a hierarchical deep clustering objective to organize surface activities into a hierarchy. We evaluate the learned hierarchy using expert annotations on two separate 4D datasets of sandy beach sites and their combination. Our results show that DTC with GeoMorphograms achieves the highest agreement with expert judgment at the taxonomy level comprising eight major process types ($F_1=0.78$, match accuracy $=0.92$), outperforming notably dimensionality reduction and conventional flat clustering, which are recent alternative methods applied to process similar time series data. The learned taxonomy separates major erosion- and deposition-dominated activities and further distinguishes finer subtypes with interpretable characteristics related to change magnitude, duration, compactness, and temporal evolution. These results demonstrate that DTC provides a scalable and interpretable route from 4D change detection to a data-driven, expert-supported surface process taxonomy, thereby advancing automated knowledge derivation and organization for understanding surface dynamics in topographic monitoring.

\end{abstract}

%% Keywords

\end{frontmatter}

% \linenumbers

\section{Introduction}
% ABT AND - BUT - Therefore
% Terminology:
% topographic change - measured elevation diff
% 4D-OBC / surface activity - extracted sample
% surface process type - learned/interpreted category
% surface process taxonomy - final hierarchical structure

Topographic landscapes are typically shaped by numerous surface processes that often operate under natural and anthropogenic forcing across a wide range of spatial and temporal scales \citep{deboerHierarchiesSpatialScale1992a,huggettFundamentalsGeomorphology2016}. On sandy beaches, for example, storms trigger rapid erosion, waves and tides redistribute sediment, wind produces localized deposition and erosion patterns, and human activities modify the surface through intervention or recreational use \citep{kuschnerusCoastalChangePatterns2021,kuschnerusIdentifyingTopographicChanges2024a,hulskemperCoastalProcessUnderstanding2026}. Most of these processes are not directly observable. Instead, they have to be inferred from measured topographic change, that is, spatiotemporal elevation differences derived from repeated surface observations. Many existent studies focus on detecting specific processes or landforms, such as rockfalls, landslides, and soil erosion \citep{williamsOptimising4DSurface2018,kromerAutomatedTerrestrialLaser2017,eltnerSoilMicrotopographyChange2018}. However, increasingly long-term monitoring offers a broader systematic perspective, in which the full range of observed surface dynamics is organized into interpretable surface process types. Such a perspective is necessary not only for describing individual processes but also for comparing recurring activities, identifying dominant process mechanisms, and ultimately understanding interactions among different surface processes. The key challenge is therefore to move beyond the measurement of topographic change and toward the derivation of a surface process taxonomy from spatiotemporal topographic data \citep{eitel3DNewSpectrum2016,woodcockTransitioningChangeDetection2020}.

Modern LiDAR techniques provide an effective basis for acquiring high-resolution 3D information of topographic surfaces across natural environments \citep{irwinLightDetectionRanging2025}. In particular, permanent laser scanning (PLS) systems can now generate dense 4D point clouds at hourly to seasonal timescales \citep{eitel3DNewSpectrum2016,lindenberghPermanentTerrestrialLaser2025}. Such high-frequency 4D data create new opportunities for environmental monitoring of coastal areas, rivers, landslides, glaciers, and other dynamically evolving landscapes \citep{kromerAutomatedTerrestrialLaser2017,williamsOptimising4DSurface2018,odeaContinuousCoastalMonitoring2019,woodcockTransitioningChangeDetection2020}. Recent studies have investigated the detection and quantification of topographic change from 4D point clouds. Methods such as robust co-registration \citep{yangSupervoxelbasedTargetlessRegistration2023,yangPiecewiseICPEfficientRobust2025a}, uncertainty-aware distance estimation \citep{lagueAccurate3DComparison2013b,winiwarterM3C2EPPushingLimits2021}, and object-based spatiotemporal segmentation \citep{anders4DObjectsbychangeSpatiotemporal2020,andersFullyAutomaticSpatiotemporal2021} have advanced the reliable detection of spatiotemporal surface activities from 4D point clouds. In particular, 4D objects-by-change (4D-OBCs) provide an object-based approach for extracting and analyzing individual surface activities. These advances have enabled the detection of where and when surface activities occur and the extraction of large numbers of surface activities from point cloud time series. Nevertheless, they represent an unordered collection of potentially large varieties of surface process types.

For large-scale and long-term monitoring, as demonstrated for sandy beach environments \citep{vosHighresolution4DTerrestrial2022,vos4DSpatiotemporalLaser2023a}, \citet{anders4DObjectsbychangeSpatiotemporal2020} detect thousands of 4D-OBCs. To support geomorphic analysis and environmental management, a successive step requires organizing these surface activities into a meaningful structure that provides an overview of recurring surface process types. This step is similar to deriving a taxonomy, which provides an organized system for comparing complex instances, identifying common types, and describing variation within and between groups \citep{godfrayChallengesTaxonomy2002}. A surface process taxonomy would allow surface activities to be systematically compared across space and time, support the identification of dominant and rare process types, and provide a basis for post-hoc analysis of process interactions and environmental drivers. However, deriving such a taxonomy for surface processes remains challenging because benchmark labeled datasets are not available, the number of surface process types is unknown a priori, and can vary between geographic settings. At the same time, surface activities often vary substantially in spatial extent, duration, geometry, magnitude, and internal dynamics \citep{wangUnsupervisedDeepClustering2025}. 

Recent studies have addressed this interpretation gap by combining 4D-OBCs with descriptive features and unsupervised clustering. For example, handcrafted descriptors derived from segmented 4D-OBCs have been used with Self-Organizing Maps (SOMs) to group similar surface activities and support their interpretation in relation to potential driving forces, such as wind and tidal records \citep{hulskemperUncoveringCausationShortTerm2026,hulskemperCoastalProcessUnderstanding2026}. Related workflows based on handcrafted descriptors and static clustering have also provided useful summaries of coastal change patterns \citep{kuschnerusCoastalChangePatterns2021,hulskemperCharacterizationMorphologicalSurface2022}. These studies demonstrate that 4D-OBCs are meaningful units for surface activity analysis and that clustering analysis can support the interpretation of surface process types. However, the representation of each surface activity still depends on manually designed features and requires domain knowledge, typically restricted to a single geographic setting. This limits scalability and flexibility, in particular when the relevant process properties are not known in advance. These current challenges motivate us to develop a reverse engineering approach that can summarize and extract domain knowledge, i.e., process properties, from the observed surface activities. A data-driven approach is therefore needed to learn process-relevant representations directly from detected 4D-OBCs and to organize them into a hierarchical structure without predefined labels or prior knowledge of the number of surface process types.

To address this, we introduce DeepTopoClustering (DTC), an unsupervised framework for deriving hierarchical surface process taxonomies from 4D point cloud data at the level of individual surface activities. The framework first represents each detected 4D-OBC as a so-called GeoMorphogram, which we develop as a distributional sequence describing the temporal evolution of topographic change within a local region (i.e., the spatial extent of a 4D-OBC). Based on these GeoMorphograms, we adapt a deep clustering-based model, following \citet{mautzDeepECTDeepEmbedded2020}, to jointly learn latent embeddings and a hierarchical clustering tree that separates activities into groups from coarse to fine levels of detail. We apply the proposed method to real-world 4D datasets acquired from two separate sandy beaches \citep{vosHighresolution4DTerrestrial2022,vos4DSpatiotemporalLaser2023a}, where several thousand 4D-OBCs were automatically extracted \citep{andersFullyAutomaticSpatiotemporal2021,hulskemperUncoveringCausationShortTerm2026}. The resulting surface process taxonomy is evaluated using expert annotations that quantify the agreement between learned clusters and human interpretation. The main contributions of this research are the following:

\begin{itemize}
    \item We present an end-to-end, object-centric pipeline that transforms 4D point clouds into an interpretable, hierarchical taxonomy of surface process types, addressing the gap between 4D topographic change detection and surface process understanding.
    \item We introduce GeoMorphograms as standardized representations for modeling individual 4D-OBCs as comparable distributional sequences, enabling discrimination of surface activities while accounting for irregular geometric properties, varying duration, and inherent noise within 4D-OBCs.
    \item We develop DeepTopoClustering as an unsupervised hierarchical deep clustering framework that derives surface activity clusters without requiring known labels or a predefined number of clusters.
    \item We derive the first data-driven process taxonomy of sandy beach surface processes from 4D point cloud data.
    \item We present an expert evaluation strategy for validating hierarchical clustering results and determining an optimal number of clusters in the absence of ground-truth labels.
\end{itemize}

\section{Related work}
The advanced technical capabilities of acquiring multi-temporal point clouds from LiDAR and photogrammetry have enabled near-continuous observation of surface dynamics. This has led to a shift from static 3D analysis to 4D point cloud time series, where topographic information is observed repeatedly over time at high frequency with respect to the observed phenomena \citep{eitel3DNewSpectrum2016,qin3DChangeDetection2016,woodcockTransitioningChangeDetection2020,lindenberghPermanentTerrestrialLaser2025}. Under such paradigm transition, recent research has developed along a connected direction: automatically acquiring 3D data near-continuously over time and aligning them in a common coordinate reference system, detecting and quantifying significant topographic change with associated uncertainty between epochs, and interpreting change information in terms of underlying surface processes \citep{lindenberghPermanentTerrestrialLaser2025}. While the first two steps are extensively developed, the transition from change detection to process-level understanding remains limited. In the following, we review existing work along this direction.

\paragraph{\textbf{Change detection and quantification}}

Quantifying topographic change between 3D data acquired at two successive epochs is a central topic in topographic monitoring \citep{qin3DChangeDetection2016}. A standard approach is Multiscale Model-to-Model Cloud Comparison (M3C2) \citep{lagueAccurate3DComparison2013b}, which estimates a spatially averaged distance directly between bitemporal point clouds along locally determined surface normals. By avoiding the transformation of the unstructured point cloud into a mesh or grid, M3C2 provides robust change estimates in rough topography and heterogeneous sampling under local planarity assumptions. Several variants have been proposed to improve accuracy and adapt M3C2 to specific surface conditions, e.g., correspondence-driven plane-based M3C2 \citep{zahsCorrespondencedrivenPlanebasedM3C22022a} and M3C2 with error propagation \citep{winiwarterM3C2EPPushingLimits2021}.

Beyond bitemporal comparison, the full temporal dimension of 4D datasets has been incorporated into change analysis. Kalman filtering \citep{winiwarterFullFourdimensionalChange2023b} is used to account for spatiotemporal uncertainties, reduce noise, and handle missing data across multiple epochs. To jointly exploit spatial and temporal coherence, object-centric change analysis has been introduced with the 4D objects-by-change (4D-OBCs) method \citep{anders4DObjectsbychangeSpatiotemporal2020}. 4D-OBCs delineate the spatial and temporal extent of surface activities by grouping spatially connected regions with similar temporal behavior. Each 4D-OBC is automatically extracted from 4D point clouds and represents a coherent surface activity that evolves over time, without predefined temporal windows \citep{andersFullyAutomaticSpatiotemporal2021}. This allows surface activity to be analyzed at the occurrence level rather than over the full point cloud time series. Despite these advancements, most existing methods focus on accurately estimating \textit{where}, \textit{when}, and \textit{how much} topographic change occurs. They provide a necessary foundation for 4D topographic monitoring, but do not directly support higher-level interpretation in terms of which process type the detected surface activities may represent.

\paragraph{\textbf{From change analysis to characterization}}
% 4D change analysis and activity objects
% Change detection (where/what magnitude) --> Change characterization (what kind of activity objects) --> Knowledge discovery (taxonomy, hierarchies).

Building on topographic change quantification in near-continuous time series, the nature of topographic changes can be characterized with respect to temporal behavior. A straightforward approach is to apply clustering algorithms to identify temporal patterns. For example, different clustering methods, such as k-means, agglomerative clustering, and DBSCAN, have been applied to derive meaningful patterns from point cloud time series for sandy beaches \citep{kuschnerusCoastalChangePatterns2021}, past-glacial valleys \citep{wangInvestigationRiverbankPeat2025}, vegetation \citep{puttonenClusteringFrameworkMonitoring2019}, and landslides \citep{weidnerAlgorithmMeasuringLandslide2021}. These studies show that unsupervised clustering can derive recurring change patterns, but they are commonly applied to point-wise or grid-wise full time series and therefore remain tied to fixed spatial resolutions rather than spatiotemporally variable surface activities.

Object-centric approaches address this limitation by using 4D Objects-by-Change (4D-OBCs) as natural units for describing surface activities, related to erosion, deposition, and redistribution processes \citep{anders4DObjectsbychangeSpatiotemporal2020}. Because 4D-OBCs vary in spatial extent, duration, and internal geometry, direct comparison between activities is difficult for machine learning methods that require standardized inputs. Recent research has addressed this by representing each 4D-OBC using handcrafted descriptors and resampled time series representations. Temporal descriptors, such as duration and timing of change magnitude, and geometric descriptors, such as location, area, and volume change, can be derived to summarize each surface activity. Time series include seed time series \citep{hulskemperCharacterizationMorphologicalSurface2022} or multivariate statistical time series  \citep{hulskemperCoastalProcessUnderstanding2026}, to enable clustering or classification. While these representations have shown value for interpreting geomorphic phenomena, they also introduce limitations: single time series are location-dependent and do not capture the full spatial variability of the 4D-OBC; data gaps and noise can reduce robustness; and resampling may cause information loss about the temporal shape of the activities. In addition, handcrafted descriptors depend on prior assumptions about which properties are relevant, while a generalized definition of process types is not available, even when considering domain expertise, which is limited to known processes. Handcrafted descriptors may overlook complex or unexpected patterns that are not known to analysts in advance. Overall, existing characterization workflows remain limited in their ability to compare diverse surface activities and to derive a systematic surface process taxonomy from long-term 4D topographic monitoring data.

\paragraph{\textbf{Deep clustering for unsupervised class derivation}}
% Clustering --> DeepECT

Recent advances in deep learning have shown strong potential for learning representations and discovering patterns from high-dimensional data, including spatiotemporal data, in an unsupervised manner \citep{bengioRepresentationLearningReview2013,wangDeepLearningSpatioTemporal2022}. In particular, deep clustering methods jointly learn latent embeddings and cluster assignments, enabling data-driven automatic clustering without requiring labeled data as input. Representative approaches include DeepCluster \citep{caronDeepClusteringUnsupervised2018}, which alternates between k-means clustering on extracted features and updating the network using pseudo-labels, and Deep Embedded Clustering (DEC; \citealp{xieUnsupervisedDeepEmbedding2016a}), which refines latent embeddings through iterative clustering with soft-assignments. Several variants of deep clustering methods have demonstrated effectiveness in image and 3D data analysis, especially when labeled data are limited or expensive to obtain \citep{zhouComprehensiveSurveyDeep2024,degelisDC3DCDUnsupervisedLearning2023,liuPointClusterDeepClustering2024}.

In remote sensing, similar concepts have been explored to address the lack of annotated datasets in real-world scenes. For example, DeepCropClustering \citep{wangDeepCropClusteringDeepUnsupervised2025} combines contrastive learning and deep clustering using nearest-farthest neighbors to derive different crop types from satellite imagery. DC3DCD \citep{degelisDC3DCDUnsupervisedLearning2023} applies the DeepCluster principle to 3D change detection, resulting in different change types by manually mapping pseudo-labels derived from bitemporal point clouds. In its first application to multi-temporal 3D change analysis, DEC was applied to time series derived from 4D-OBCs to characterize surface activities within erosion and deposition as two separate categories \citep{wangUnsupervisedDeepClustering2025}. Existent work indicates that deep clustering can extract useful latent features from spatiotemporal data and support change analysis without manual labels. However, most existing approaches use flat clustering and require the number of clusters to be predefined. This is a major limitation for deriving surface process types, where their number is usually unknown a prior and where process categories may be more adequately represented as a hierarchy than as flat clusters. In a hierarchy, higher levels may share the same drivers and overall types; lower levels may be separated by fine-grained properties.

To address these limitations, recent methods have explored hierarchical or flexible clustering structures that do not require a fixed number of clusters in advance. Examples include DeepECT \citep{mautzDeepECTDeepEmbedded2020}, deep taxonomic networks \citep{wangDeepTaxonomicNetworks2025}, and DeepDPM \citep{ronenDeepDPMDeepClustering2022}. Such approaches are particularly relevant for comprehensive topographic monitoring where the underlying category structure is unknown, ambiguous, or multi-level rather than being specifically defined by analysts. However, no existing framework integrates object-centric 4D change analysis, representation learning, and hierarchical deep clustering to derive generic interpretable surface process types. This motivates us to develop an unsupervised deep clustering approach to learn representations of surface activities and organize them into a structured taxonomy without prior assumptions on the number or properties of process types.

\section{Data and methods}

\begin{figure*}[!htb]
\begin{center}
    \includegraphics[width=0.85\textwidth]{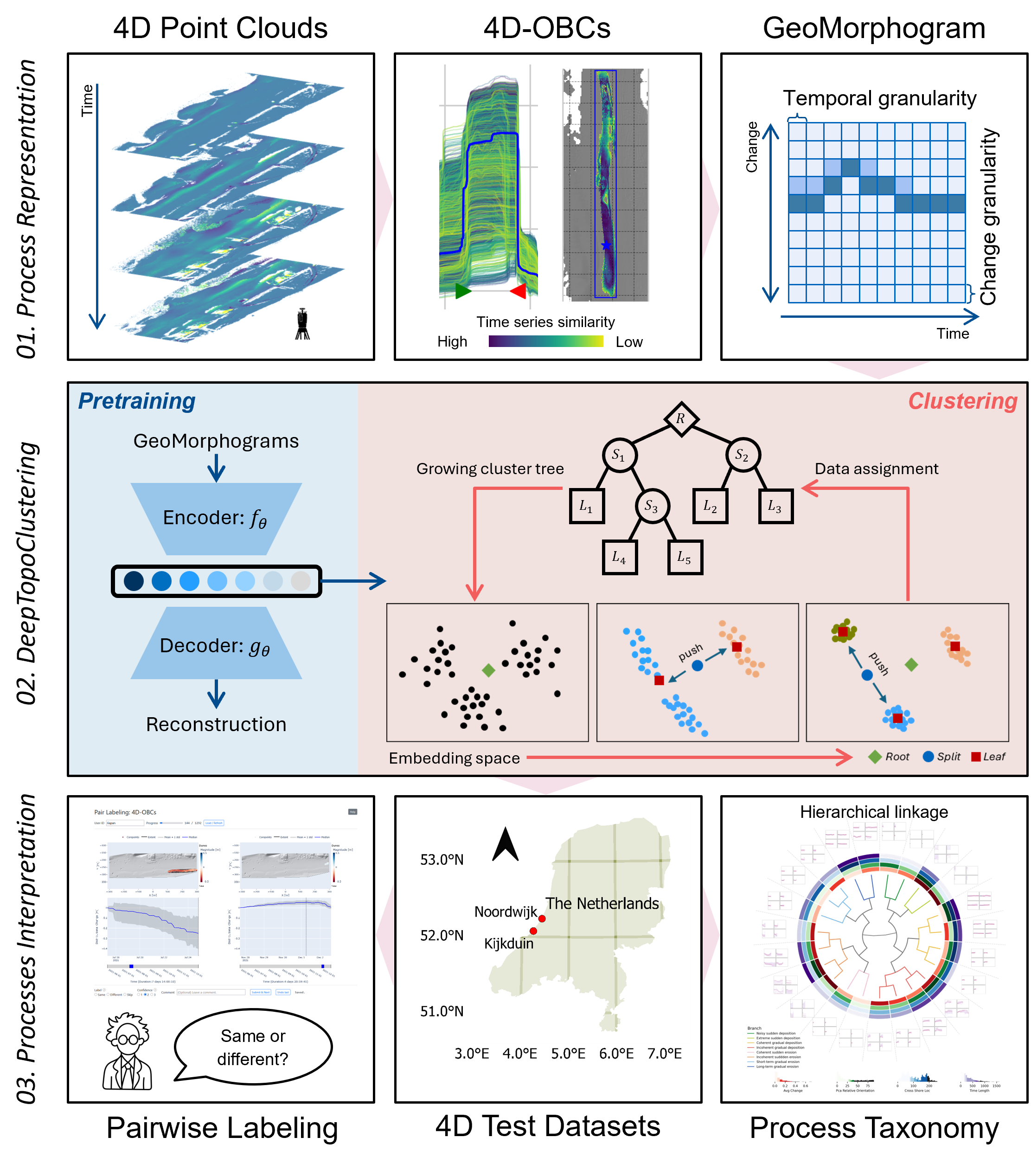}
	\caption{Overall research framework for deriving surface process taxonomy from 4D point clouds, including representing extracted 4D surface activities as GeoMorphograms, organizing them into learnable hierarchical clusters, and validating the learned taxonomy on 4D test datasets with expert annotations.} 
\label{fig:method-overview}
\end{center}
\end{figure*}

We develop a three-stage framework for deriving a data-driven surface process taxonomy from near-continuous 4D point clouds. The first stage is \textbf{Process Representation}: surface activities are extracted as 4D objects-by-change (4D-OBCs), which define discrete local surface activities evolving over time. Each 4D-OBC is then represented as a GeoMorphogram (Section~\ref{sec:geomorphogram}), a distributional sequence that describes the temporal evolution of topographic change within a local region. The second stage is \textbf{DeepTopoClustering}: these GeoMorphograms are embedded into a compact feature space during a pretraining phase and are subsequently clustered using DTC (Section~\ref{sec:deeptopoclustering}), a hierarchical deep clustering approach that organizes similar surface activities into a divisive tree structure and reveals multiple levels of process similarity. The third stage is \textbf{Process Interpretation}: the learned hierarchical linkage is evaluated on 4D test datasets (Section~\ref{sec:data}) using expert annotations to quantitatively assess and qualitatively interpret the derived surface process taxonomy (Section~\ref{sec:expert-labeling}). The main steps of our framework are illustrated in Figure~\ref{fig:method-overview}, and details on data and methods are outlined in the following subsections.

\subsection{Data and preprocessing}\label{sec:data}

\begin{figure*}[!htb]
\begin{center}
    \includegraphics[width=0.9\textwidth]{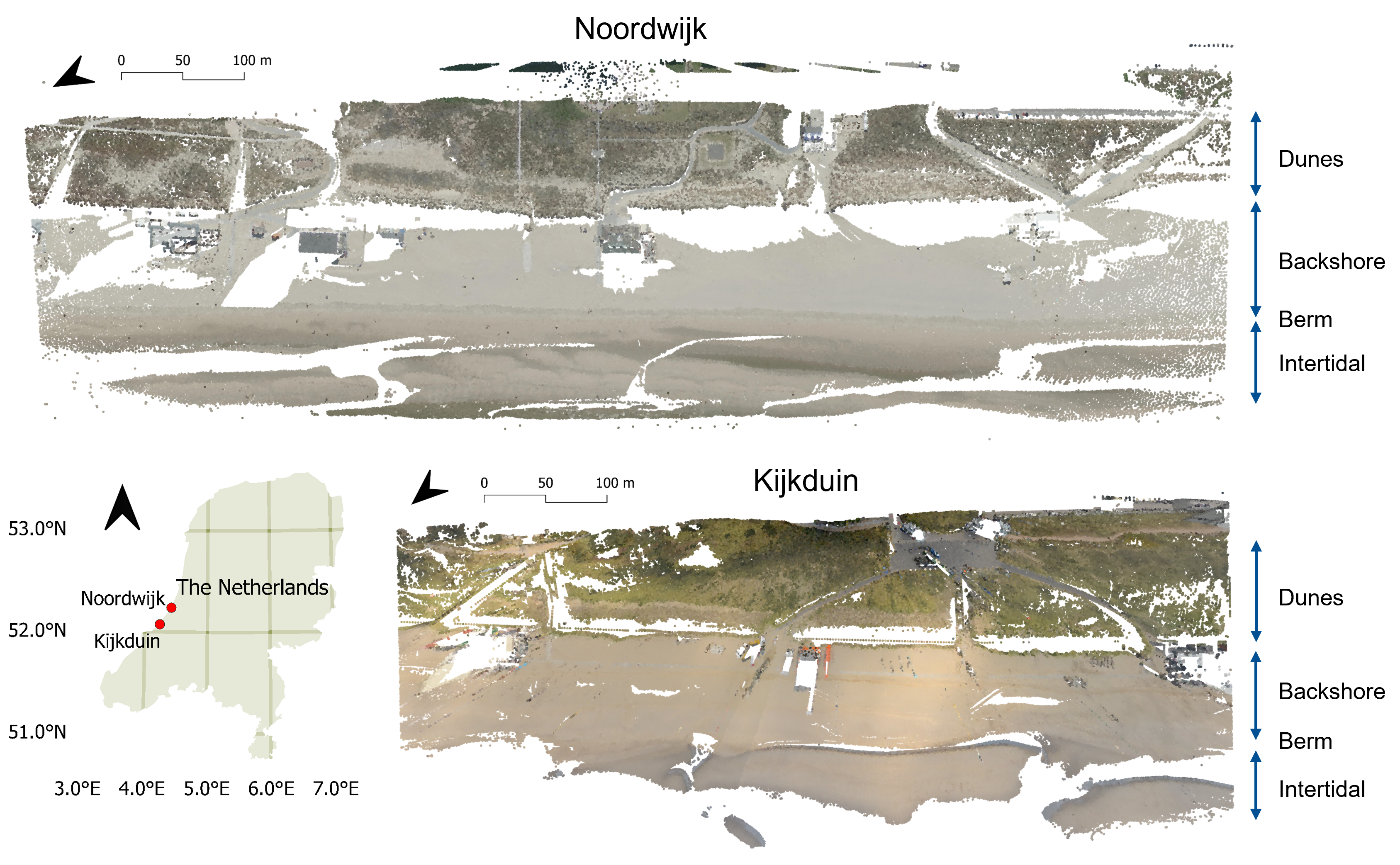}
	\caption{Permanent laser scanning point cloud of the study areas in Noordwijk and Kijkduin sandy beach. The terrestrial laser scans are colorized using high-resolution aerial imagery in 2021 (Noordwijk) and in 2016 (Kijkduin). The aerial imagery and administrative areas are provided by the Public Service On the Map (\href{https://www.pdok.nl/}{PDOK}) in the Netherlands.} % \href{https://www.pdok.nl/-/nu-hoge-resolutie-luchtfoto-2024-bij-pdok}{WMTS}
\label{fig:data-map}
\end{center}
\end{figure*}

For our study, we use two 4D point cloud datasets acquired with PLS systems installed at the sandy beach of Kijkduin \citep{vosHighresolution4DTerrestrial2022} and Noordwijk \citep{vos4DSpatiotemporalLaser2023a} in the Netherlands (Figure~\ref{fig:data-map}). At each site, a Riegl VZ2000 laser scanner was mounted on a stable frame on a hotel roof overlooking the study areas. Kijkduin (52.07°N, 4.22°E) was monitored during the winter and spring of 2016 - 2017 for approximately six months, and Noordwijk (52.24°N, 4.42°E) was monitored between July 2019 and July 2022 for about three years. Hourly laser scans were collected over the sandy beach, including its dunes, backshore, berm, and intertidal zones, resulting in thousands to tens of thousands of point cloud epochs. Details on the monitoring setup and data characteristics are provided in \citet{vosHighresolution4DTerrestrial2022} and \citet{vos4DSpatiotemporalLaser2023a}, and a summary is presented in Table~\ref{tab:dataset}. The resulting near-continuous point cloud time series are subsequently processed to derive spatiotemporal elevation change and extract discrete activity objects as 4D-OBCs.

\begin{table}[!htb]
\caption{Data summary of the two open source PLS datasets (i.e., Kijkduin and Noordwijk sandy beaches) used in this research.}
\begin{adjustbox}{width=1.0\columnwidth}
\begin{tabular}{@{}lll@{}}
\toprule
Dataset        & Kijkduin   (NL)        & Noordwijk   (NL)        \\ \midrule
Location       & 52.07°N,4.22°E         & 52.24°N, 4.42°E         \\
Extent         & $300\times600$ m        &  $350\times1000$ m            \\
Time interval & 1   hour               & 1   hour                \\
Time range     & Nov. 2016 – May 2017 & July 2019 – July 2022 \\
Epochs         & 2,942                 & 21,194                  \\ \bottomrule
\end{tabular}
\end{adjustbox}
\label{tab:dataset}
\end{table}

The PLS datasets are processed through several typical steps, including co-registration, change quantification, and surface activity extraction \citep{lindenberghPermanentTerrestrialLaser2025}. We use the resulting 4D-OBCs extracted in previous work from the Kijkduin \citep{andersFullyAutomaticSpatiotemporal2021} and Noordwijk \citep{hulskemperCoastalProcessUnderstanding2026} 4D datasets. Each 4D-OBC represents an individual surface activity derived from the topographic change (here, M3C2 distances \citep{lagueAccurate3DComparison2013b}) and serves as a fundamental unit for subsequent representation learning and clustering. In total, three datasets are considered: the Kijkduin dataset with 2,021 4D-OBCs, the Noordwijk dataset with 4,654 4D-OBCs, and a merged sandy beach activities dataset, referred to as BeachOBC, with 6,675 4D-OBCs (i.e., the sum of both). Each dataset is partitioned by time of occurrence: all 4D-OBCs are grouped by month using the middle of their time span, and the training and unseen subsets are randomly split at an 80-20 ratio within each month. This strategy ensures that the model is trained to capture full process types over the entire monitoring period and evaluated on previously unseen surface activities.

\subsection{GeoMorphogram: a distributional sequence representation of a surface activity}\label{sec:geomorphogram}
% Problem - limitation - design - definition - intuition

To represent the full surface activity rather than selected summaries from a single time series, we introduce the \textit{GeoMorphogram}, a distributional sequence representation that describes the temporal evolution of topographic change within each surface activity. A GeoMorphogram represents a 4D-OBC as an ordered sequence of topographic change distributions over time. At each epoch, topographic change values are extracted within the spatial extent of the 4D-OBC and aggregated into a histogram. This produces a 2-dimensional representation in which the horizontal axis denotes time and the vertical axis denotes topographic change magnitude. In contrast to time series summaries based on mean, median, or other object-level statistics, the GeoMorphogram preserves the distributional state of a surface activity at each epoch, including both the magnitude and spread of change.

\begin{figure*}[!htb]
\begin{center}
    \includegraphics[width=0.9\textwidth]{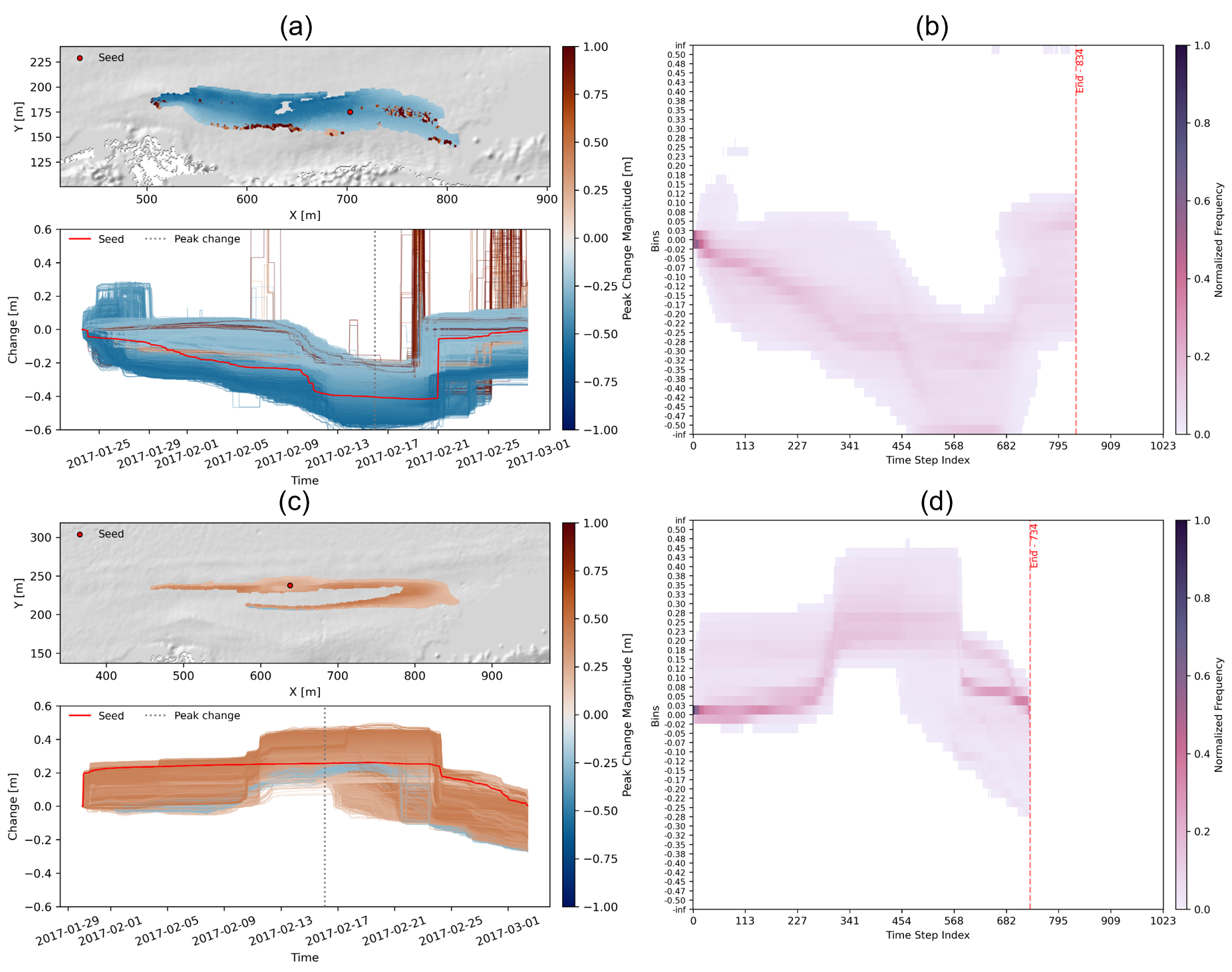}
	\caption{Examples of 4D objects-by-change (4D-OBCs) and their corresponding GeoMorphograms. Subplots (a) and (c) show erosion- and deposition-dominated 4D-OBCs, respectively. For each object, the upper map shows the spatial extent with core points colored by their peak change magnitude and the seed point marked in red. The lower plot shows the change time series of all core points within the object, colored by their peak change magnitude; the red line indicates the seed time series and the gray dotted line marks the peak change epoch. Subplots (b) and (d) show the corresponding GeoMorphograms for the erosion 4D-OBC (a) and the deposition 4D-OBC (c), respectively. Color intensities show the distribution of change values at each epoch, and the activity duration is marked using a red dashed vertical line.}
\label{fig:4d-obcs}
\end{center}
\end{figure*}

To construct a GeoMorphogram, let $O_i$ denote a 4D-OBC, representing a surface activity occurring within spatial extent $\Omega_i \in \mathbb{R}^2$ and temporal duration $T_i=\{t_1,\dots,t_D\}$. For each epoch $t_d$, the topographic change within the 4D-OBC is represented by a set of topographic change values, measured relative to the surface state at the first epoch $t_1$, at points $p$:
\begin{equation*}
    \{\Delta z(p,t_d) \mid p \in \Omega_i \}.
\end{equation*}

A histogram function $\mathcal{H}(\cdot)$ is then applied to aggregate these values into a discrete distribution over $B$ bins:
\begin{equation*}
    h_i(t_d) = \mathcal{H}(\{\Delta z(p,t_d) \mid p \in \Omega_i \}) \in \mathbb{R}^B;
\end{equation*}
The histogram is normalized to represent a probability distribution:
\begin{equation*}
    \sum^B_{b=1} h_i^{(b)}(t_d) = 1.
\end{equation*}
This object-wise normalization reduces the influence of spatial variability, such as differences in object size, because each epoch is represented by relative frequencies rather than absolute point counts. By encoding frequency distributions instead of raw topographic change magnitudes, the GeoMorphogram emphasizes change states and their temporal evolution rather than absolute topographic differences. This facilitates consistent comparison among surface activities with different spatial scales and geometries.

The GeoMorphogram of object $O_i$ is then defined as the ordered sequence of the topographic change distribution over time duration $D$:
\begin{equation}
    \mathbf{G}_i = [h_i(t_1), \dots, h_i(t_D)] \in \mathbb{R}^{B \times D}
\end{equation}
where each element represents the occurrence frequency of a given change magnitude at a specific epoch. GeoMorphograms are constructed using fixed discretization parameters to ensure comparability across samples. In this study, the vertical axis is discretized using a bin resolution of 0.025~m over the range $[-0.5~\text{m}, 0.5~\text{m}]$, based on the 95\% range ($[-0.25~\text{m},0.36~\text{m}]$) of topographic change values in the datasets. The defined boundaries cover the major range of change values with a margin while keeping GeoMorphograms not too large, and the bin resolution is determined with respect to the minimum detectable change for the datasets \citep{kuschnerusStatisticallyAssessingVertical2024}. Two additional bins are included at both ends to capture values outside this range, resulting in $B = 42$ bins. This configuration balances the need to retain meaningful variation in topographic change magnitude with consideration of measurement noise. The horizontal time axis is defined from the start of each activity and discretized at an hourly resolution up to a maximum duration of $D = 1024$~hours. This duration covers 97\% (904 at 95\%) of all 4D-OBCs while limiting excessive padding. 4D-OBCs shorter than this maximum duration are zero-padded at the end, whereas longer samples, which account for 3\% of the dataset, are downsampled to 1024 epochs to preserve long-term deformation behavior. All GeoMorphograms are translated relative to the initial state of each 4D-OBC and represented as normalized probability distributions. These choices yield GeoMorphograms with consistent dimensions, enabling direct comparison among surface activities with different spatial extents and temporal durations.

The resulting GeoMorphogram encodes both temporal dynamics and distributional variability within each 4D-OBC. For example, Figure~\ref{fig:4d-obcs} shows that an erosion-dominated activity (a) produces temporal distributions (b) that are consistently skewed toward negative topographic changes, whereas a deposition-dominated activity (c) generates temporal distributions (d) concentrated in positive change bins. Although most time series broadly follow the trend of the seed time series, the GeoMorphograms also reveal wider or bimodal distributions across epochs. This reflects that sediment may be redistributed in a complex manner even within a single 4D-OBC.

\subsection{DeepTopoClustering for hierarchical grouping}\label{sec:deeptopoclustering}
We use the Geomorphograms as a standardized representation of surface activities and map them into a latent space, where activities with similar temporal-distributional change patterns are expected to have similar embeddings. To learn this representation in an unsupervised manner, the latent embeddings are first learned through self-reconstruction using an autoencoder (AE) \citep{bengioRepresentationLearningReview2013}. Because the number of surface process types is also unknown a priori, the clustering framework should not require a predefined number of clusters and should allow the clustering structure to evolve during training. We therefore develop DeepTopoClustering (DTC), a hierarchical deep clustering framework adapted from the Deep Embedded Clustering Tree (DeepECT) method \citep{mautzDeepECTDeepEmbedded2020} for spatiotemporal topographic change analysis.

DTC consists of two main components: (1) a feature extraction module that learns latent embeddings from GeoMorphograms, and (2) a hierarchical clustering module that organizes these embeddings into a binary tree through iterative node splitting. In contrast to a flat clustering method, this structure allows broad differences among surface activities to be represented at shallow tree levels and finer variations to be represented at deeper levels. In our framework, DeepECT is adapted to operate on GeoMorphograms, to jointly optimize representation learning and hierarchical clustering, and to decode node centers into GeoMorphogram prototypes for interpretation. The learning process iteratively refines both the embedding space and the tree structure, enabling surface activities to be grouped at multiple levels of detail.

\paragraph{\textbf{Feature extractor}}
We design an autoencoder-based feature extractor consisting of a four-layer convolutional neural network (CNN) as the encoder $f_\theta(\cdot)$, and a mirrored transposed CNN as the decoder $g_\theta(\cdot)$. This design treats the GeoMorphogram as a structured temporal-distributional image rather than a time series representation, allowing the model to capture local patterns in both temporal and spatial dimensions of topographic changes. The learnable parameters $\theta$ are randomly initialized and optimized during training.

Given a GeoMorphogram $\textbf{G}_i \in \mathbb{R}^{B \times D}$, the encoder maps the input into a latent embedding with a fixed dimension of 16:
\begin{equation}
    z_i = f_\theta(\textbf{G}_i).
\end{equation}
The decoder then reconstructs the input from the embedding: 
\begin{equation}
    \tilde{\textbf{G}}_i = g_\theta(z_i).
\end{equation}
Because GeoMorphograms represent normalized distribution sequences and may contain padded epochs beyond the valid duration of a surface activity, reconstruction is optimized using a masked Kullback–Leibler (KL) divergence loss. Let $h_i^{(b)}(t_j)$ and $\tilde{h}_i^{(b)}(t_j)$ denote the true and the reconstructed probabilities for bin $b$ at time $t_j$, respectively. The reconstruction loss is defined as
\begin{equation} 
    \mathcal{L}_{REC} = \frac{1}{|\mathcal{S}|} \sum_{i=1}^\mathcal{S} [ \frac{1}{m_i} \sum_{j=1}^{m_i}\sum_{b=1}^{B}h_i^{(b)}(t_j)\log(\frac{h_i^{(b)}(t_j)}{\tilde{h}_i^{(b)}(t_j)})],
    \label{eq:loss-rec}
\end{equation}
where $m_i \leq D$ is the valid duration of object $i$, $D$ is the defined maximum duration across all objects, $B$ is the number of topographic change bins, and $\mathcal{S}$ is the set of training samples. The masking ensures that only valid epochs contribute to the reconstruction objective. Minimizing the reconstruction loss $\mathcal{L}_{REC}$ encourages the latent space to preserve the main temporal-distributional patterns of surface activities while reducing the dimensionality of the input. This pre-training phase aims to establish a meaningful autoencoder and feature space for subsequent hierarchical clustering, as shown in stage 2 of Figure~\ref{fig:method-overview}.

\paragraph{\textbf{Hierarchical clustering}}
Given the latent embeddings, we implement the clustering module following the hierarchical clustering strategy of DeepECT. It constructs a binary tree, where each node is either a leaf node or a split node. The split node has two child nodes, which form a sibling pair and represent two subclusters of the samples assigned to their parent. Each node is associated with a vector $\mu_n$ in the latent space, which serves as the node center. The tree is initialized with a single root node containing all samples and grows progressively by splitting heterogeneous leaf nodes during optimization.

While the tree is growing, the leaf node with the highest internal variability is selected to split. Internal variability is measured by the sum of squared distances between the node center and the embeddings assigned to that node. The selected node is divided into two child nodes by applying k-means++ with $k=2$ to the assigned embeddings. We use k-means++ as it is considered more accurate and faster than standard k-means \citep{arthurKmeansAdvantagesCareful2007}. This split operation introduces a new local separation in the latent space while preserving the previously learned hierarchical structure.

The clustering objective encourages samples assigned to each node to be compact along the direction that separates the node from its sibling. As demonstrated by \citet{mautzDeepECTDeepEmbedded2020}, directly minimizing the Euclidean distance between samples and node centers can suppress structural information that is orthogonal to the separation direction. DeepECT, therefore, defines a sibling-specific projection direction. For two sibling nodes with centers $\mu_n$ and $\mu_m$, the normalized direction from node $m$ to node $n$ is defined as
\begin{equation}
\rho_n = \frac{\mu_n - \mu_m}{\lVert \mu_n - \mu_m \rVert_2}
\end{equation}
where $\rho_n$ defines the principal direction along which the sibling nodes are separated. The data compression (DC) loss then penalizes deviations of the assigned embeddings from their corresponding node centers along this direction:
\begin{equation}
\mathcal{L}_{DC} \;=\; \frac{1}{|\mathcal{N}|\cdot|\mathcal{S}|}\sum_{n\in\mathcal{N}}\sum_{\textbf{G}_i\in\mathcal{S}}
\big|\rho_n^{\top}\big(\mu_n-f_\theta(\textbf{G}_i)\big)\big|
\label{eq:loss-dc}
\end{equation}
where $\mathcal{N}$ is the set of non-root nodes and $\mathcal{S}$ the set of samples. This loss compresses the embeddings within each node along the sibling-separation direction, thereby preserving variation that may be relevant for deeper splits. As a result, strong surface activity differences can be separated at shallow levels of the tree, whereas finer differences can emerge at deeper levels.

Intuitively, during the optimization, the hierarchy is learned in a top-down manner through an alternating process of compression, splitting, and pruning. After a certain number of training epochs, the node with the most heterogeneous samples is selected as the split candidate, which creates two child nodes and builds a new local separation direction. Nodes that contain very few samples are pruned to avoid unstable or unnecessary over-split. Through this adaptive process, the tree grows progressively and organizes surface activities into a hierarchy that captures multiple levels of separation. For our dataset, we update the tree structure every 3 training epochs and prune nodes containing fewer than 10 samples.

\paragraph{\textbf{Node prototyping}}
Since the node centers are learnable parameters and the latent space changes during training, we adopt the node center (NC) loss proposed in DeepECT \citep{mautzDeepECTDeepEmbedded2020} to stabilize the clustering structure throughout the full optimization process. The NC loss updates the vector of each leaf node by aligning it with the mean embedding of data samples assigned to that node. Formally, the node center loss is defined as:
\begin{equation}
\mathcal{L}_{NC} = \frac{1}{|\mathcal{N}|}\sum_{n\in\mathcal{N}}\frac{1}{|\mathcal{S}_n|}\sum_{\textbf{G}_i\in\mathcal{S}_n}\|\mu_n - f_\theta(\textbf{G}_i)\|_2
\label{eq:loss-nc}
\end{equation}
where $\mathcal{N}$ denotes the set of nodes, $\mathcal{S}_n$ is the set of samples assigned to node $n$, and $\mu_n$ represents the cluster centroid. During optimization, the latent embeddings $f_\theta(\textbf{G}_i)$ are treated as constant when updating the node centers, such that the loss $\mathcal{L}_{NC}$ only pulls the node center towards the mean of its assigned samples rather than pulling the samples toward the center. Furthermore, the split node centers are not directly updated, but are derived by the weighted average over child nodes. 

Importantly, in our framework, each node center $\mu_n$ serves as a latent prototype of an embedding group. By passing this prototype through the decoder $g_\theta$, we obtain a reconstructed GeoMorphogram $\tilde{\textbf{G}}_n = g_\theta(\mu_n).$ that represents the characteristic topographic change pattern of the surface activity cluster. These decoded prototypes can provide an interpretable summary of each cluster and allow direct visualization and comparison of the various surface activities. The learned hierarchy functions as both a grouping structure in the latent space and a set of representative topographic change patterns that support the interpretation of complex surface activities themselves.

\paragraph{\textbf{Loss computation}}
The overall DTC objective combines the reconstruction loss (Eq.~\ref{eq:loss-rec}), the data compression loss (Eq.~\ref{eq:loss-dc}), and the node center loss (Eq.~\ref{eq:loss-dc}) in a weighted manner:
\begin{equation}
    \mathcal{L} = \lambda_{REC}\mathcal{L}_{REC} + \lambda_{DC}\mathcal{L}_{DC} + \lambda_{NC}\mathcal{L}_{NC}
    \label{eq:loss-all}
\end{equation}
where $\lambda_{REC}, \lambda_{DC},\lambda_{NC}$ are weighting coefficients that balance the scale of representation learning and hierarchical clustering.

Training is performed in two stages. First, the autoencoder is pretrained for 20 training epochs using only the reconstruction loss to learn a latent space that captures the temporal-distributional patterns of GeoMorphograms. Second, hierarchical clustering is introduced by jointly optimizing the combined loss in Eq.~\ref{eq:loss-all}. During joint optimization, embeddings are extracted from all samples, samples are assigned to the nearest leaf nodes in the latent space using Euclidean distance, and the tree is updated based on assignments according to the splitting and pruning criteria. The whole autoencoder and node centers are then optimized with the combined loss. We stop training when the tree reaches a maximum of 50 leaf nodes or when the number of nodes remains stable after a split iteration. This process enables DTC to progressively derive a hierarchical structure of surface activity patterns without requiring the number of process types to be specified in advance.

All implementation parameters are fixed across the Kijkduin, Noordwijk, and combined BeachOBC experiments unless stated otherwise. We use a batch size of 64 and apply a dropout rate of 0.1 to reduce overfitting. Optimization is performed using the AdamW optimizer \citep{loshchilovDecoupledWeightDecay2019}. A learning rate of $2\times10^{-3}$ with weight decay $1\times10^{-4}$ is used during pretraining to accelerate convergence, followed by a reduced learning rate of $8\times10^{-4}$ with weight decay $5\times10^{-5}$ during joint optimization to ensure stable refinement of both embeddings and cluster structure.

The loss weights are selected using Optuna \citep{akibaOptunaNextgenerationHyperparameter2019}, which performs automated optimization based on a given evaluation objective. Here, the objective is to maximize the $F1$ score computed from expert annotations on the validation set and apply the same setting on the test set, as described in Section~\ref{sec:expert-labeling}. Because our objective is to derive a general process taxonomy, here for sandy beach surface activities, we use the combined BeachOBC to decide the final hyperparameter setting, which is then used for the individual Kijkduin and Noordwijk datasets. The resulting loss weights are $\lambda_{REC}=1.0$, $\lambda_{DC}=3.79$, $\lambda_{NC}=7.63$, which provide a balanced contribution of reconstruction, data compression, and node center objectives.

\subsection{Evaluation}\label{sec:expert-labeling}
\begin{figure*}[!htb]
\begin{center}
    \includegraphics[width=1.0\textwidth]{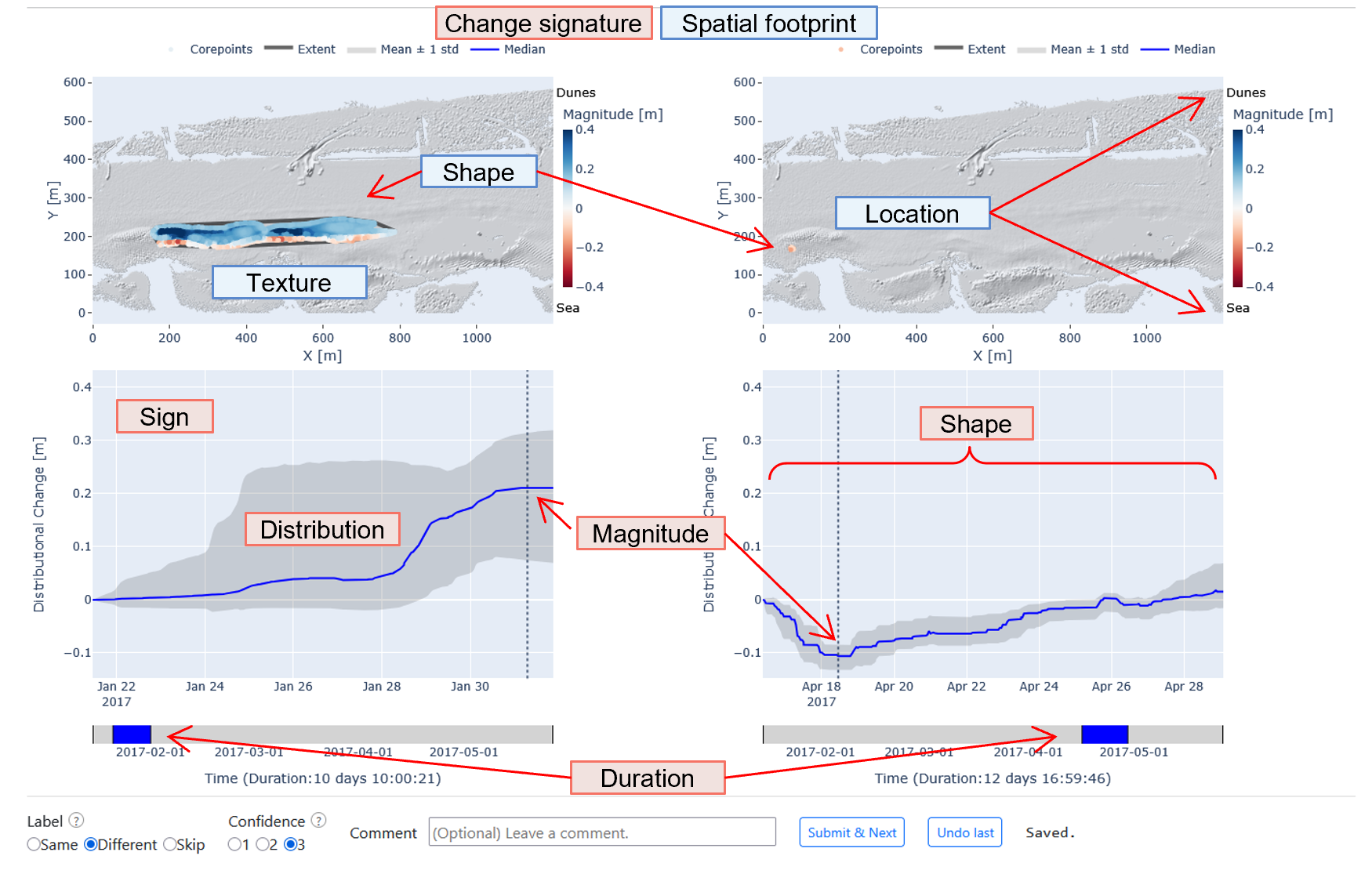}
	\caption{Labeling interface used to collect expert annotations. Each task presents a pair of 4D objects-by-change (4D-OBCs) through their spatial and temporal representation. Experts assess whether the two 4D-OBCs represent the same or different types of surface activity and provide a confidence score. The labeling interface figure is shown as guidance, as provided in the tutorial for expert annotators.}
\label{fig:labeling}
\end{center}
\end{figure*}

\paragraph{\textbf{Pairwise expert annotation}} To address the lack of labeled data and of a priori knowledge about the occurrence of surface activity types, we design a pairwise evaluation strategy that leverages expert knowledge while avoiding exhaustive object-level labeling. We select reference 4D-OBCs as a random subset and use handcrafted features to identify nearest neighbors as pairs for each 4D-OBC. The handcrafted features describe surface activities in terms of their spatial, temporal, and change-related properties, including size, shape, orientation, cross-shore location, total change volume, duration of occurrence, initial height, and initial timestamp, following \citet{hulskemperCoastalProcessUnderstanding2026}. Experts in coastal geomorphology and 4D point cloud analysis are then asked to annotate each sample pair with a binary label indicating whether two 4D-OBCs belong to the same or different types of surface activity through an online labeling interface (Figure~\ref{fig:labeling}). The final label of each pair is determined by \textit{weighted majority voting} from different experts based on annotation confidences. After assembling annotations and filtering out low-confidence labels, we obtain 467 labeled pairs out of a total of 8,754 annotations by 7 experts. We randomly select 50\% of these labeled pairs as a validation set to optimize the hyperparameters and use the other 50\% as a test set to evaluate the model's performance.

\paragraph{\textbf{Multi-level comparison}} Given that the proposed DTC framework produces a hierarchical clustering of topographic change objects, we evaluate the results against expert annotations across different levels of detail through the order of split nodes. Specifically, we consider pairs that are labeled as similar and grouped within the same node by DTC as true positives (TP), and pairs labeled differently and assigned to separate nodes as true negatives (TN). Pairs labeled as different but grouped in the same node are false positives (FP), and pairs labeled as similar but separated by DTC are false negatives (FN). This evaluation reflects the nature of the hierarchical clustering: clearly distinct surface activities are expected to separate at shallow levels, whereas more subtle distinctions may emerge only at deeper levels. Conversely, pairs labeled as similar should remain in the same cluster until finer levels of the tree. This evaluation protocol does not aim to define a single ``correct'' cluster, but rather to quantify the consistency between data-driven clustering and expert interpretation. In doing so, it provides a practical and interpretable way to validate the uncovered structure and to indicate meaningful patterns that underlie observed surface activities. Furthermore, it enables identifying the level of granularity at which the learned hierarchy best aligns with expert-annotated distinctions of topographic surface activities.

\paragraph{\textbf{Evaluation metrics}} To quantify our model's performance, we calculate precision, recall, F1 score, and match accuracy based on the binary agreement between clustering results and expert labels using the following formulas:
\begin{align*}
    Precision &= \frac{TP}{TP+FP}\\
    Recall &= \frac{TP}{TP+FN}\\
    F_1 &= 2\times \frac{Precision \times Recall}{Precision+Recall}\\
    Match\ Accuracy &= \frac{TP+TN}{TP+FP+FN+TN}.\\
\end{align*}

\subsection{Deriving a surface process taxonomy}\label{sec:deriving-taxonomy}

To interpret the learned hierarchy, we visualize the samples and cluster-wise characteristics of each leaf node, inspired by the hierarchical morphotope classification of built form introduced by \citet{fleischmannHierarchicalMorphotopeClassification2026}. To identify the main surface process types, we do not manually select a preferred level of detail. Instead, we follow the pairwise evaluation described in Section~\ref{sec:expert-labeling} and select the clustering level that shows the highest agreement with expert annotations. The number of nodes at this level is reported as an expert-supported number of clusters $k$. We therefore interpret the selected clusters as data-driven, expert-supported, and interpretable surface process types. This strategy allows the full taxonomy to be derived from the learned hierarchy, while the reported main types remain constrained by expert judgment.

To characterize clusters, we first compute selected object-wise handcrafted features, including average change, PCA relative orientation, cross-shore location, and time duration of activity, to showcase physical, geometry, domain, and temporal properties. Average change is calculated as the absolute total change volume ($m^3$) divided by the footprint area ($m^2$), resulting in an area-normalized change magnitude in meters. PCA relative orientation describes the shape orientation of the 4D-OBC footprint. It is defined as the acute angle (0°--90°) between the principal axis of the point distribution, estimated by PCA, and the waterline direction, where 0° indicates an orientation parallel to the waterline and 90° indicates an orientation perpendicular to it. Cross-shore location (m) is defined as the perpendicular distance between the median coordinates and the waterline boundary. Activity time duration (h) is defined as the number of active epochs of each 4D-OBC. These object-wise features provide descriptive context for the learned clusters and support the interpretation of the representative surface process types in Section \ref{sec:results}.

\subsection{Competing methods}

To assess the respective contributions of the input representation and the clustering method, we compare DTC with several baselines. Principal component analysis (PCA) + k-means is used as a linear dimensional reduction baseline. Autoencoder (AE) + k-means and AE + agglomerative clustering evaluate whether AE-based embeddings improve clustering when the representation and clustering steps are optimized separately. DEC \citep{xieUnsupervisedDeepEmbedding2016a} is included as a flat deep clustering baseline, following \citet{wangUnsupervisedDeepClustering2025}, which jointly optimizes representation learning and clustering, but requires a predefined number of target clusters. For input representations, we compare our proposed GeoMorphograms with the mean time series of 4D-OBCs, which are comparable to the inputs used in existing work \citep{wangUnsupervisedDeepClustering2025,hulskemperCharacterizationMorphologicalSurface2022,hulskemperCoastalProcessUnderstanding2026}. This setup allows us to evaluate the effects of input representation, latent embedding learning, and hierarchical clustering separately.

All experiments in this study are conducted on a Linux workstation (Ubuntu 22.04.5 LTS) equipped with an Intel® Xeon(R) w7-3455 CPU, 128 GB of RAM, and an NVIDIA RTX A4500 GPU (20 GB). This setup is used for both training and inference of the models.

\section{Results}\label{sec:results}

\subsection{Quantitative performance}\label{sec:performance}

\begin{figure}[!htb]
\begin{center}
    \includegraphics[width=1.0\columnwidth]{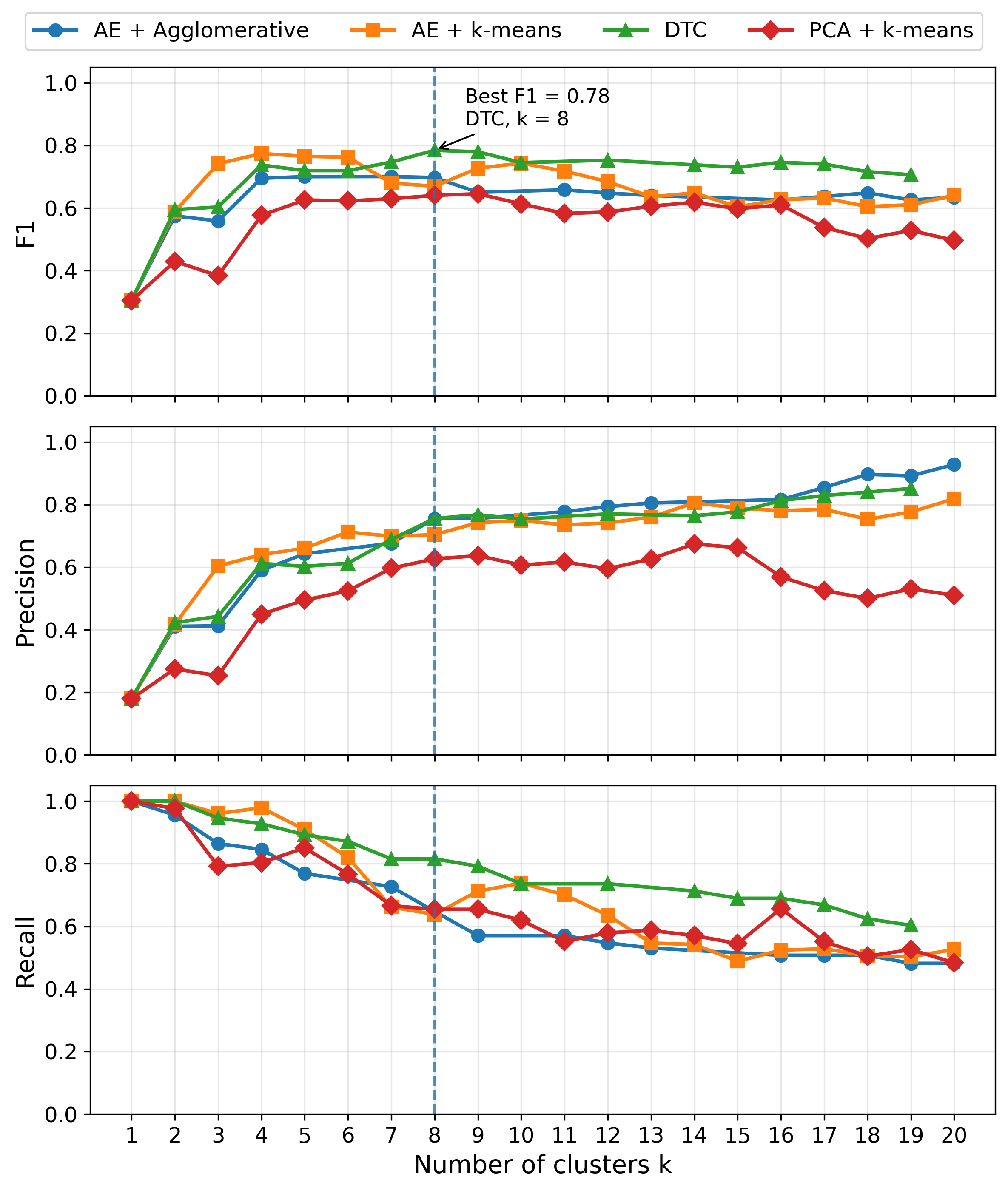}
	\caption{Evaluation metrics computed on pairwise annotations across different numbers of clusters $k$ on the test set. The dashed line marks the selected level, $k=8$, where the highest overall F1 score is achieved by our approach.}
\label{fig:k-selection-test}
\end{center}
\end{figure}

\begin{figure*}[!htbp]
\begin{center}
    \includegraphics[width=1.0\textwidth]{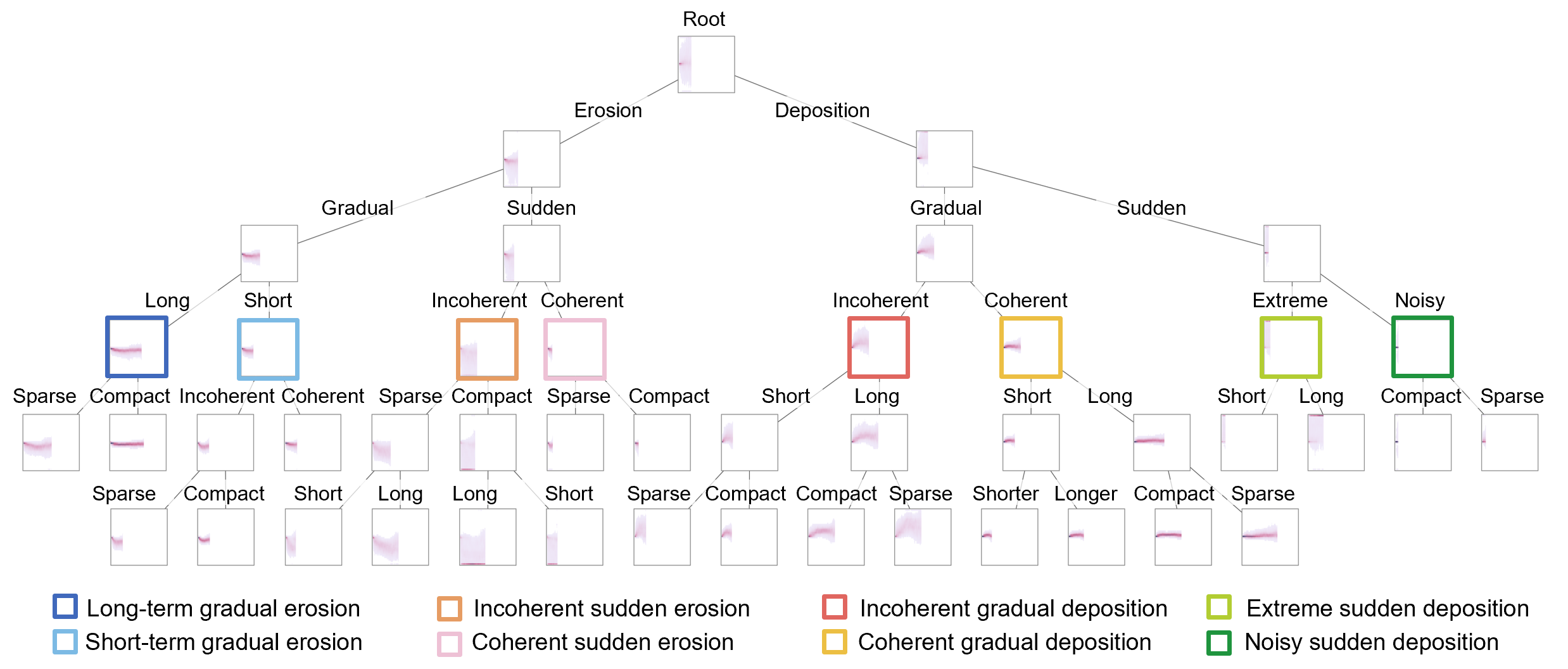}
	\caption{Learned hierarchical tree from the BeachOBC dataset. Each node image represents the decoded GeoMorphogram prototype from the corresponding cluster center, and each edge indicates the parent--child relation between nodes. Text labels indicate the dominant visual differences between sibling branches and are used to support the interpretation of the splitting sequence. The derived expert-supported types (at $k=8$) are highlighted by colorful boxes, and the branches are named based on the distinguishing properties along the tree.}
\label{fig:prototype}
\end{center}
\end{figure*}

\begin{figure*}[!htb]
\begin{center}
    \includegraphics[width=1.0\textwidth]{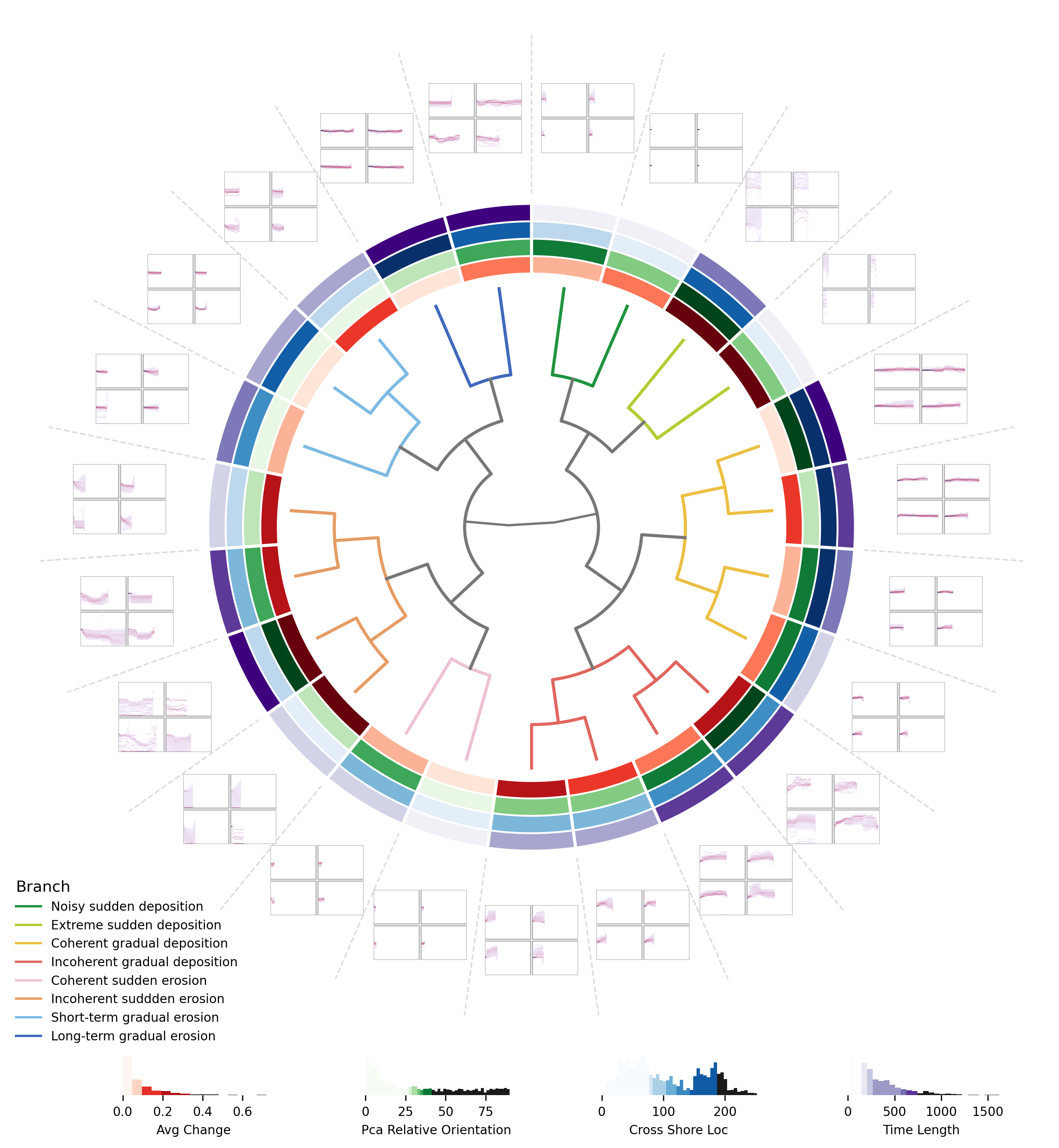}
	\caption{Surface process taxonomy derived from the BeachOBC dataset, consisting of combined 4D-OBCs detected from the Kijkduin and Noordwijk datasets. The circular tree shows the selected hierarchical structure at $k=8$, the outer panels show representative GeoMorphograms for each leaf node, and the surrounding feature rings summarize cluster-wise physical properties.}
\label{fig:taxonomy}
\end{center}
\end{figure*}

\begin{figure}[!htb]
\begin{center}
    \includegraphics[width=1.0\columnwidth]{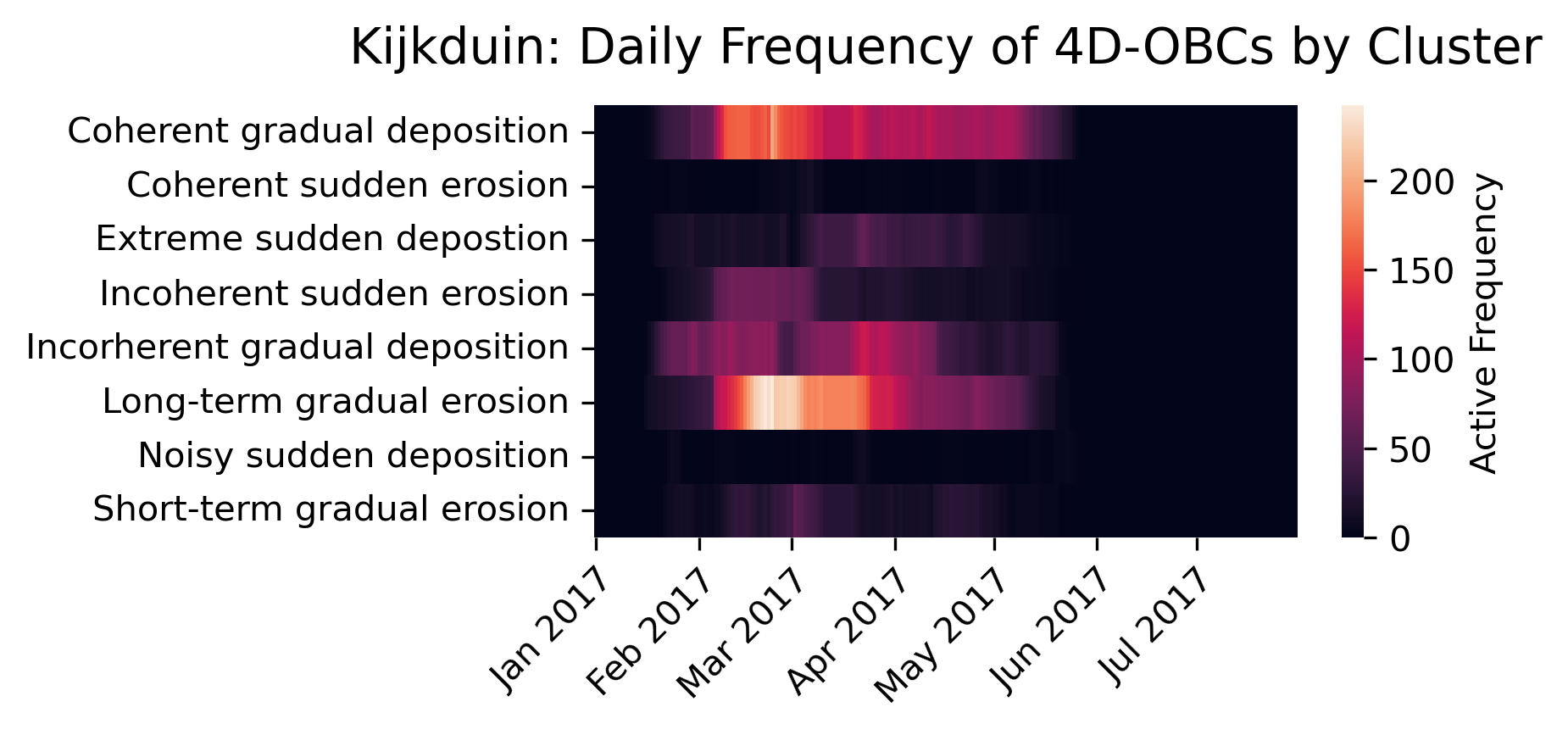}
	\caption{Temporal frequency of different surface process clusters at Kijkduin. Frequency is computed as the number of active 4D-OBCs per day for each selected cluster.}
\label{fig:kijkduin_freq}
\end{center}
\end{figure}

\begin{figure*}[!htb]
\begin{center}
    \includegraphics[width=1.0\textwidth]{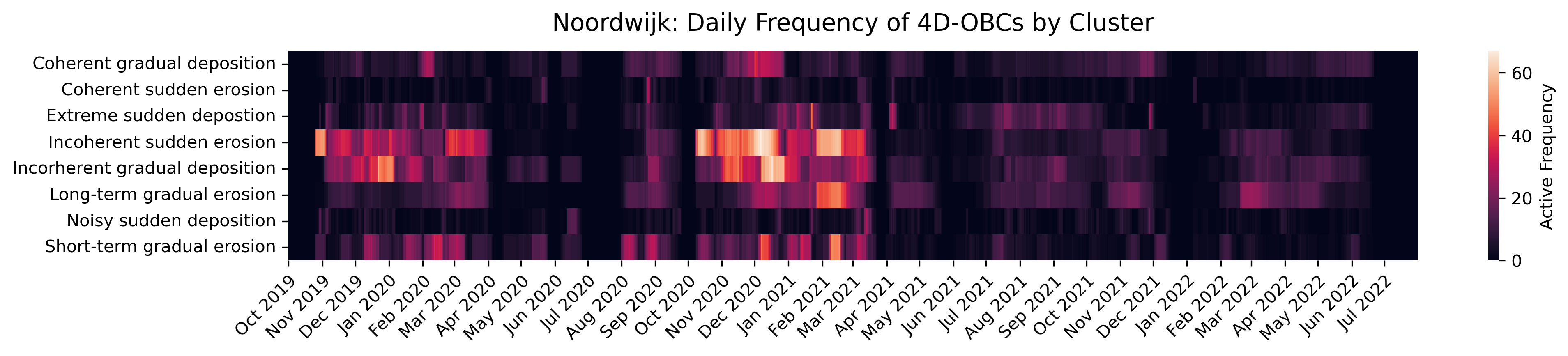}
	\caption{Temporal frequency of different surface process clusters at Noordwijk. Frequency is computed as the number of active 4D-OBCs per day for each selected cluster.}
\label{fig:noord_freq}
\end{center}
\end{figure*}

\begin{table*}[!htb]
\caption{Quantitative comparison of different clustering methods and input representations on the BeachOBC, Kijkduin, and Noordwijk datasets. The selected number of clusters (k = 8) is determined from the DTC hierarchy using the highest validation F1 score. Best results for each dataset and metric are shown in bold. Precision (P), recall (R), F1, and match accuracy (Acc) are computed from test sets of expert pairwise annotations.}
\begin{adjustbox}{width=1.0\textwidth}
\begin{tabular}{llllllllllllllll}
\hline
Method             & Input          & \multicolumn{4}{c}{BeachOBC}                                           & \multicolumn{1}{c}{} & \multicolumn{4}{c}{Kijkduin}                                  & \multicolumn{1}{c}{} & \multicolumn{4}{c}{Noordwijk}                                 \\ \cline{3-16} 
                   &                & P             & R             & \multicolumn{1}{c}{F1} & Acc           &                      & P             & R             & F1            & Acc           &                      & P             & R             & F1            & Acc           \\ \hline
PCA~+~k-means      & Time series    & 0.26          & \textbf{0.88} & 0.40                   & 0.53          &                      & 0.26          & \textbf{0.83} & 0.39          & 0.56          &                      & 0.33          & \textbf{0.84} & 0.48          & 0.66          \\
AE + k-means       & Time series    & 0.57          & 0.77          & 0.66                   & 0.86          &                      & 0.59          & 0.68          & 0.63          & 0.86          &                      & 0.61          & 0.79          & 0.69          & 0.87          \\
AE + Agglomerative & Time series    & 0.55          & 0.71          & 0.62                   & 0.84          &                      & 0.50          & 0.69          & 0.58          & 0.83          &                      & 0.63          & 0.67          & 0.59          & 0.84          \\
DEC                & Time series    & 0.53          & 0.67          & 0.59                   & 0.83          &                      & 0.76          & 0.69          & 0.72          & 0.91          &                      & 0.72          & 0.73          & 0.72          & 0.90          \\
DTC                & Time series    & 0.54          & 0.68          & 0.60                   & 0.84          &                      & 0.51          & 0.64          & 0.57          & 0.84          &                      & 0.76          & 0.74          & 0.75          & 0.91          \\ \hline
PCA~+~k-means      & GeoMorphograms & 0.63          & 0.65          & 0.64                   & 0.87          &                      & 0.63          & 0.74          & 0.68          & 0.88          & \textbf{}            & 0.61          & 0.64          & 0.63          & 0.86          \\
AE + k-means       & GeoMorphograms & 0.70          & 0.64          & 0.67                   & 0.89          &                      & 0.77          & 0.52          & 0.62          & 0.89          &                      & 0.77          & 0.73          & 0.75          & 0.91          \\
AE + Agglomerative & GeoMorphograms & \textbf{0.76} & 0.65          & 0.70                   & 0.90          & \textbf{}            & 0.49          & 0.67          & 0.57          & 0.82          &                      & 0.80          & 0.69          & 0.75          & 0.91          \\
DEC                & GeoMorphograms & 0.74          & 0.70          & 0.72                   & 0.90          &                      & 0.67          & 0.45          & 0.53          & 0.87          &                      & 0.80          & 0.79          & 0.79          & 0.92          \\
DTC                & GeoMorphograms & \textbf{0.76} & 0.82          & \textbf{0.78}          & \textbf{0.92} & \textbf{}            & \textbf{0.92} & \textbf{0.83} & \textbf{0.87} & \textbf{0.95} &                      & \textbf{0.84} & 0.82          & \textbf{0.83} & \textbf{0.94} \\ \hline
\end{tabular}
\label{tab:results}
\end{adjustbox}
\end{table*}

Figure~\ref{fig:k-selection-test} shows the evaluation metrics obtained for different numbers of clusters on the BeachOBC test set. The metric plot for the validation set is provided in the appendix (Figure~\ref{fig:k-selection-valid}). The highest validation and test F1 scores are obtained with DTC at $k=8$, where precision and recall are best balanced: smaller values of $k$ tend to over-aggregate surface activities that are different according to expert annotations, whereas larger values potentially over-split similar activities. We therefore select $k=8$ as the expert-supported target number of clusters of major surface process types and for comparison with flat clustering baselines, which are evaluated using the same number of clusters.

Table~\ref{tab:results} reports the quantitative results on the BeachOBC, Kijkduin, and Noordwijk datasets. Overall, DTC with GeoMorphograms as input achieves the highest F1 score, precision, and match accuracy across all three datasets and obtains the best or near-best recall performance. On the combined BeachOBC dataset, DTC achieves an F1 score of 0.78 and an accuracy of 0.92, outperforming all competing methods. Similar results are observed on the individual datasets, where DTC reaches an F1 score of 0.87 on Kijkduin and 0.83 on Noordwijk. 

Across most methods, GeoMorphogram inputs yield higher F1 scores than mean time series inputs. For PCA~+~k-means, the F1 score increases from 0.40 to 0.64 on BeachOBC, from 0.39 to 0.68 on Kijkduin, and from 0.48 to 0.63 on Noordwijk. For DTC, the F1 score increases from 0.60 to 0.78 on BeachOBC, from 0.57 to 0.87 on Kijkduin, and from 0.75 to 0.83 on Noordwijk. These results indicate that GeoMorphograms provide a more discriminative representation of object-wise surface activity than mean time series. The comparison between clustering methods also shows that the best performance is obtained when GeoMorphograms are combined with the hierarchical DTC model.

The precision and recall values show different performance patterns across methods. PCA~+~k-means with time series input shows high recall on BeachOBC (0.88), Kijkduin (0.83), and Noordwijk (0.84) but low precision on BeachOBC (0.26), Kijkduin (0.26), and Noordwijk (0.33), indicating that many similar pairs are grouped together, but many different pairs are also incorrectly assigned to the same cluster. In contrast, DTC with GeoMorphogram input achieves the second-highest or equal-highest recall on BeachOBC (0.82), Kijkduin (0.83), and Noordwijk (0.82), while also achieving the highest precision on BeachOBC (0.76), Kijkduin (0.92), and Noordwijk (0.84), suggesting that the learned hierarchy can preserve broad similarities while separating heterogeneous surface activity types more effectively.

\subsection{Learned hierarchy tree}\label{sec:tree}

\begin{figure*}[!htb]
\begin{center}
    \includegraphics[width=1.0\textwidth]{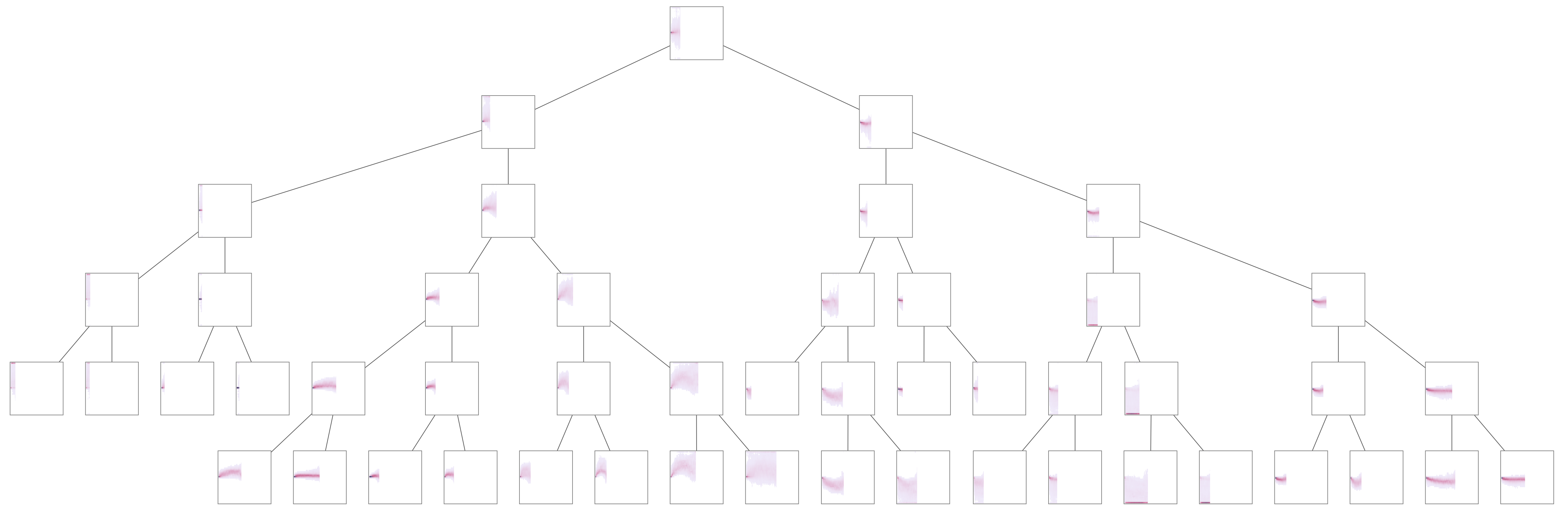}
	\caption{Learned hierarchical tree from the Noordwijk dataset. Each node image represents the decoded GeoMorphogram prototype from the corresponding cluster center, and each edge indicates the parent--child relation between nodes.}
\label{fig:prototype-noordwijk}
\end{center}
\end{figure*}

The hierarchical clustering structure learned by DTC on the BeachOBC dataset is visualized as a binary decision tree in Figure~\ref{fig:prototype}. Each node represents a cluster of surface activities in the latent space and is illustrated by decoding its learned node center into a GeoMorphogram prototype. The decoded prototypes summarize the representative temporal distribution of topographic change within the corresponding node, while the edges indicate the parent--child relations formed through the splitting process. The clustering trees learned from the individual Kijkduin and Noordwijk datasets are provided in the Appendix (Figure~\ref{fig:prototype-kijkduin} and Figure~\ref{fig:prototype-noordwijk}).

The branch labels in Figure~\ref{fig:prototype} describe visually observable properties derived from the decoded prototypes. Change direction refers to the dominant sign of relative topographic change, and change magnitude refers to the distance of the dominant change values from zero. ``Gradual'' and ``sudden'' change describe whether the dominant change develops over several consecutive time steps or is concentrated within a short interval. ``Long'' and ``short-term'' duration refer to the active time span of the surface activity. ``Compact'' and ``sparse'' distribution describe whether the change values are concentrated near the average or spread over a broader portion of the GeoMorphogram. ``Coherent'' and ``incoherent'' evolution refer to whether the main distributional structure persists across consecutive time steps or turns directions. ``Extreme'' and ``Noisy'' refer to large-magnitude irregular changes or random fluctuations without a clear temporal pattern. These terms are therefore used to annotate the distinguishing properties at each split. Furthermore, each selected leaf node can be described by the sequence of distinguishing properties that separate it from its sibling branches. This provides a consistent naming principle for the cluster: a cluster name is based on the combination of distinguishing properties along its path, as shown in Figure~\ref{fig:prototype} bottom.

At the first split, the hierarchy separates surface activities into two main branches corresponding to major types of surface activities. The decoded prototypes show that one branch is dominated by negative topographic change values, corresponding to erosion-dominated activities, whereas the other is dominated by positive topographic change values, corresponding to deposition-dominated activities. Within the erosion-dominated branch, for example, the second level separates clusters into relatively gradual and sudden surface activities. Deeper splits further distinguish activities with different durations, degrees of coherence, and compactness of distribution. Each erosion-dominated leaf node is therefore characterized by a specific combination of properties along its decision path from the root. Within the deposition-dominated branch, the second level separates gradual and sudden surface activities, where gradual deposition is represented by a more continuous increase and sudden deposition by a more abrupt signal of positive values over a short interval. Subsequent splits further distinguish deposition activities according to temporal coherence, compactness, and duration.

\subsection{Surface process taxonomy}

The derived surface process taxonomy is shown in Figure~\ref{fig:taxonomy}. Based on the quantitative evaluation in Section~\ref{sec:expert-labeling}, the tree cut at cluster number of $k=8$ achieves the highest agreement with expert annotations, with an F1 score of 0.78 on the BeachOBC dataset (Table~\ref{tab:results}). We therefore use this level of detail to define the main types of sandy beach surface processes. The selected branches are interpreted as data-driven and expert-supported surface process types. In Figure~\ref{fig:taxonomy}, the circular tree shows the hierarchical relations among all clusters, and the surrounding panels show representative GeoMorphograms for the corresponding leaf nodes.

The cluster-wise handcrafted feature summaries are presented as feature rings around the hierarchy in Figure~\ref{fig:taxonomy}. These features include average change, PCA relative orientation, cross-shore location, and time duration, as defined in Section~\ref{sec:deriving-taxonomy}. Each feature is categorized into five color intensities for direct visual comparison. The feature rings show that different clusters are associated with different combinations of physical, geometric, domain, and temporal features. Neighboring branches share similar feature profiles, whereas more distant branches show stronger differences in one or more features. Together with the learned hierarchy and representative GeoMorphograms, this taxonomy provides the descriptive basis for interpreting the hierarchical clusters as surface process types. 

\subsection{Temporal occurrence of different process types}

To showcase the applicability of the taxonomy for a deeper understanding of surface processes on sandy beaches, we examine the temporal occurrence patterns of the identified surface process types at the two study sites. Figures~\ref{fig:kijkduin_freq} and \ref{fig:noord_freq} show the daily frequency of each selected type over the monitoring period at Kijkduin and Noordwijk, respectively. The frequency is computed as the number of active 4D-OBCs per day for each type.

At Kijkduin, the frequency of the ``long-term gradual erosion'' type increases strongly from February to March (Figure~\ref{fig:kijkduin_freq}). During the same period, the ``coherent gradual deposition'' type also shows increased activity. This peak coincides with a period of bar migration and erosion as observed by \citet{vosCrossShoreIntertidalBar2020}. At Noordwijk, a similar increase is visible in February 2021 (Figure~\ref{fig:noord_freq}), with weaker increases in 2020 and 2022. These temporal occurrence plots show when different taxonomy types are active at each site and provide the basis for comparing the timing and frequency of surface activity patterns across monitoring periods.

\section{Discussion}

\begin{figure}[!htb]
\begin{center}
    \includegraphics[width=1.0\columnwidth]{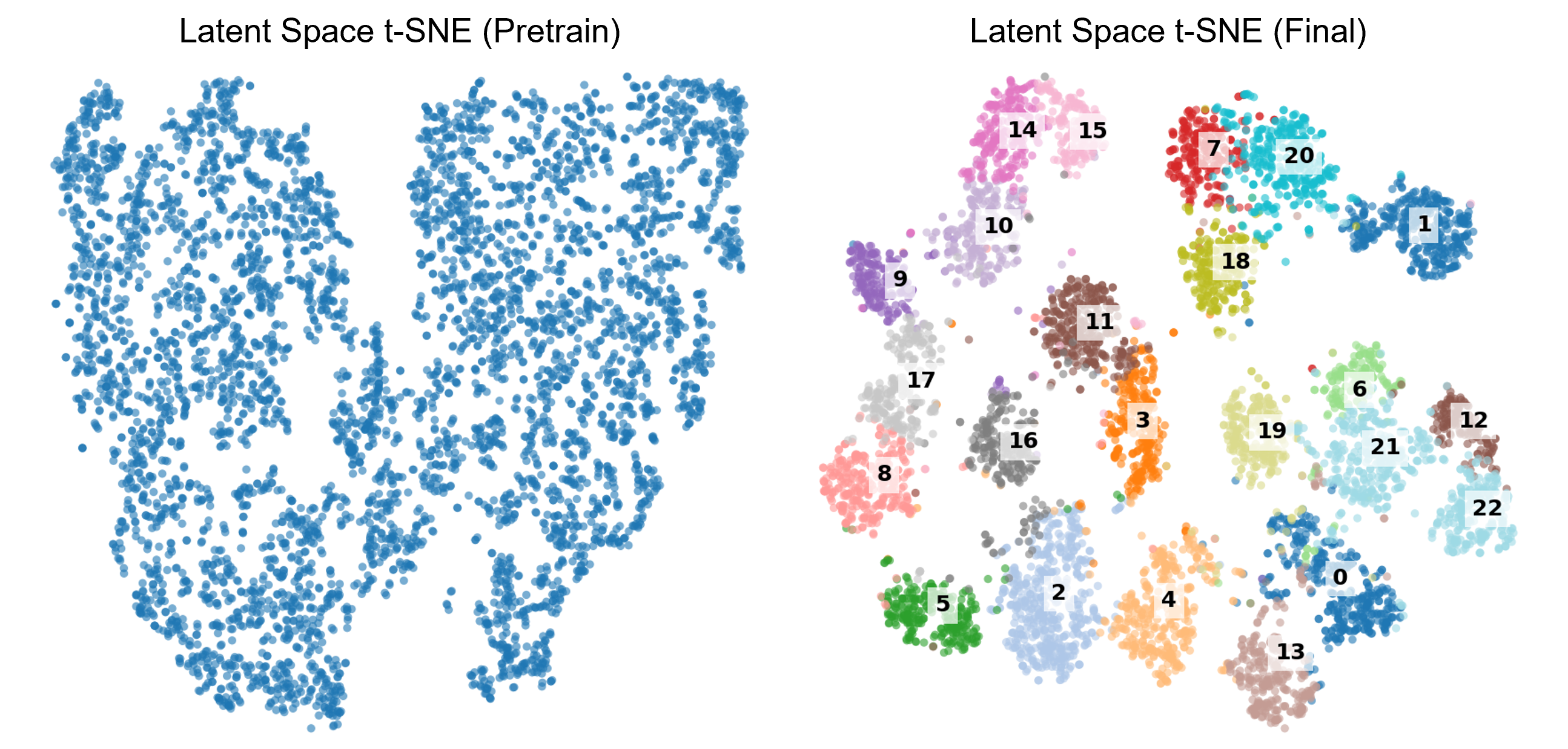}
	\caption{T-SNE visualization of the learned GeoMorphogram embeddings after pretraining (left) and clustering (right).}
\label{fig:tsne}
\end{center}
\end{figure}

\begin{figure}[!htb]
\begin{center}
    \includegraphics[width=1.0\columnwidth]{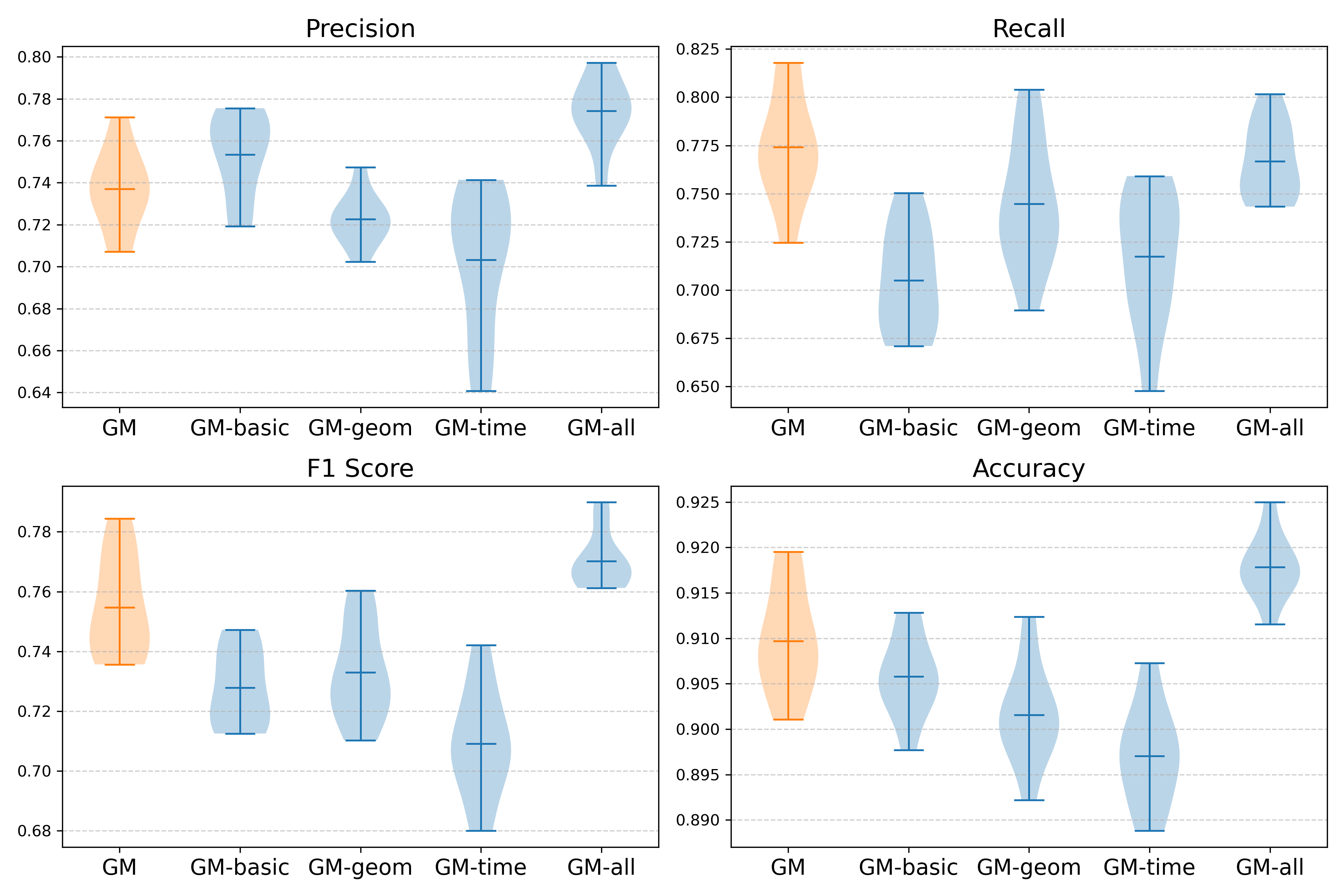}
	\caption{Performance comparison of integrating different handcrafted feature sets on the BeachOBC dataset. GM denotes the GeoMorphogram input (orange). Different conditioning settings (blue) are: (1) basic features, including area and total change volume; (2) geometric features, including area, principal axes of the point distribution estimated by PCA, PCA relative orientation, cross-shore location, initial mean elevation, and total change volume; (3) temporal features, including duration and cyclically encoded start time with sine/cosine transformation; and (4) all of the features of 1--3.}
\label{fig:conditioning}
\end{center}
\end{figure}

\subsection{GeoMorphograms and hierarchical clustering}

The quantitative results (Section~\ref{sec:performance}) highlight the importance of combining GeoMorphogram-based representation and hierarchical clustering optimization in DTC. GeoMorphograms preserve variation in change magnitude, sign, temporal coherence, and temporal evolution within each 4D-OBC. These properties are also commonly considered during expert interpretation. In contrast, mean time series summarize a 4D-OBC into a limited set of values and may therefore suppress object-wise differences relevant for distinguishing notably spatial heterogeneous process types. This explains why replacing mean time series with GeoMorphograms improves $F_1$ scores across most methods.

The precision and recall values further clarify this effect. PCA~+~k-means with time series input achieves high recall but low precision, indicating that it tends to over-aggregate surface activities. This means that most pairs identified by experts as belonging together are successfully grouped into the same cluster, but the clusters also contain many pairs that experts consider different, resulting in a large number of false-positive matches. This behavior is consistent with a time series representation that captures broad temporal trends but ignores spatial differences. In contrast, GeoMorphograms reduce this ambiguity by preserving distributionally spatial change within 4D-OBCs, leading to more balanced precision and recall, and higher F1 scores when combined with DTC.

The evolution of the embedding space provides further insight into the role of the joint optimization objective (Figure~\ref{fig:tsne}). After autoencoder pretraining, the latent embeddings exhibit a binary separation indicating erosion- and deposition-dominated surface activities. Finer distinctions between diverse clusters emerge only after the joint optimization with the hierarchical clustering objective. This demonstrates that reconstruction learning provides a meaningful initial latent space, whereas the hierarchical clustering refines the latent space towards deriving more subtle patterns beyond the primary types. These results therefore support the use of an end-to-end hierarchical deep clustering method over existing two-step workflows that separately perform feature extraction and clustering.

\subsection{Effect of integrating additional handcrafted features}

For sandy beach surface activities, object-wise handcrafted features may include area, cross-shore location, duration, magnitude, slope, and change volume \citep{hulskemperCoastalProcessUnderstanding2026}. These features are intuitive and interpretable, making them attractive for domain-specific applications. Several of them are also reflected in the cluster characteristics of the derived taxonomy, although they are not explicitly involved during the training. To examine whether additional domain knowledge can further improve DTC performance, we incorporated multiple groups of handcrafted features into the latent embeddings through a lightweight conditioning module. 

This module takes the normalized handcrafted features and uses a multi-layer perceptron (MLP) layer to project them into an 8-dimensional vector. This vector is then concatenated with the 16-dimensional GeoMorphogram embedding to form the full embeddings of surface activities. We repeat the DTC training ten times on the BeachOBC dataset using four groups of handcrafted features: (1) basic features, including area, and total change volume; (2) geometric features, including area, principal axes of the point distribution estimated by PCA, PCA relative orientation, cross-shore location, initial mean elevation, and total change volume; (3) temporal features, including duration, and cyclically encoded start time with sine/cosine transformation; and (4) all of the features of 1--3. 

The results in Figure~\ref{fig:conditioning} show that handcrafted feature conditioning provides only limited and inconsistent improvement over the GeoMorphograms alone. Conditioning with all handcrafted features yields the highest mean F1 score, precision, and accuracy, indicating that additional features can provide complementary information when combined. However, conditioning with only basic, geometric, or temporal features does not consistently improve performance and can reduce recall, accuracy, or F1 score. This suggests that the proposed GeoMorphograms already contain most of the relevant information required for clustering surface activities. Handcrafted features may provide secondary refinement, especially when multiple feature groups are combined, but simple concatenation does not essentially improve DTC's performance.

\subsection{Interpretation of the learned taxonomy}

As shown in Figure~\ref{fig:prototype}, the learned hierarchy organizes all surface activities into a hierarchical structure. Coarse levels capture broad distinctions, such as erosion-dominated and deposition-dominated activities. Deeper levels separate more specific variations in duration, change magnitude, coherence, compactness, and temporal evolution. Building such a hierarchy is label-agnostic, although deriving major surface process types still requires expert support. The selected number of clusters $k=8$ provides the highest agreement with expert annotations; however, other levels can still be useful for different analytical purposes.

The derived taxonomy (Figure~\ref{fig:taxonomy}) should therefore be interpreted as an expert-supported, data-driven organization of observed surface process types rather than as a definitive process classification. Surface processes on sandy beaches are often mixed, gradual, and context-dependent. A single 4D-OBC may contain both erosion and deposition signals, or may reflect multiple drivers acting over the same period. Therefore, process boundaries are not always discrete. The DTC provides the possibility to identify these clusters and indicates that the learned representation captures object-wise variations in surface processes that may not be explicitly recognized in existent analysis approaches.

This flexibility is important for topographic monitoring. A coastal manager may prefer coarse process types for a rapid overview, while a geomorphologist may inspect deeper branches to study subtle variations in sediment redistribution or recovery behavior. The decoded GeoMorphogram prototypes support this interpretation by showing the temporal distributional change properties of each cluster. The sequence of splits from root to leaf nodes further provides an interpretable chain of distinguishing properties, enabling systematic characterization of surface process types across multiple levels of detail.

\subsection{Applicability in the coastal domain} 

The derived taxonomy provides a comprehensive and interpretable way to summarize large volumes of topographic 4D point cloud data. Instead of inspecting thousands of individual 4D-OBCs, users can analyze the occurrence of a limited set of surface process types. This reduces the complexity of the original point cloud time series while preserving information about the specific type and timing of surface processes.

The partially overlapping occurrence patterns of similar surface process types at Kijkduin (Figure~\ref{fig:kijkduin_freq}) and Noordwijk (Figure~\ref{fig:noord_freq}), together with the relatively similar performance scores (Table~\ref{tab:results}), suggest that the derived taxonomy has potential transferability across sandy beach sites. This taxonomy can therefore be used at varying levels of detail to characterize different beaches in terms of their short-term responses through the identification of occupancy of the tree. This enables comparisons of seasonal and location-dependent variations in short-term responses of sandy beaches without any pre-defined labels.

Potentially, variations in the occupancy of taxonomy branches may also reflect differences in environmental conditions between sites and time periods. In future statistical causal analysis, this might be exploited by characterizing variations in the degree of occupancy of the tree, and linking these to time-varying environmental variables through causal inference \citep{rungeModernCausalInference2023}, so as to identify the causes of varying response of the beach system. The hierarchical structure aids in identifying at which level of detail differences start to occur at each moment in time, whereas in flat clustering, these differences might not become apparent.

Additionally, the clustering result could aid data-driven prediction of beach activities. The identified hierarchy greatly reduces the data complexity of the point cloud time series while preserving relevant change information, such that simpler one-dimensional time series based algorithms (e.g., regression, long short-term memory) can become applicable. If these are able to adequately predict future surface activities, the taxonomy, in particular, is well applicable for aiding timely response to storm erosion and response cycles through adequate beach management, as the taxonomy offers a relatively simple interpretation of expected surface activities.

\subsection{Limitations and future work}

Despite the demonstrated advantages of the proposed DTC framework, several limitations should be acknowledged. First, GeoMorphograms abstract topographic change as temporal distributions within each 4D-OBC. This makes the representation compact and comparable, but it reduces explicit spatial information. Directions of sediment transport, internal spatial structure, and interactions between neighboring activities are not directly modeled. Future work could address this limitation by incorporating spatial GeoMorphograms, graph-based interactions, or spatial attention mechanisms.

Second, our experiments are currently limited to sandy beach environments. The transferability of the learned taxonomy to other geomorphic settings, such as riverbeds, glaciers, landslides, or urban scenes, remains to be tested. These environments may contain different spatial structures, temporal scales, and process interactions. More automatic generalization of 4D-OBC extraction might support broader application of our method.

Third, the evaluation relies on expert pairwise annotations. This is practical in the absence of benchmark labels, but it cannot remove subjectivity from process interpretation. The selected level of detail is therefore best understood as the level most consistent with the available expert judgments, rather than as an absolute ground truth. This also reflects the broader challenge that surface process categories are often ambiguous and context-dependent rather than strictly defined. Semantic annotation protocols and human-in-the-loop refinement could make the evaluation more robust and make the parameterization more automatic.

\section{Conclusion}

This study addresses the challenge of deriving meaningful surface process types from PLS-acquired 4D point cloud data without predefined labels or a fixed number of target clusters. To bridge the gap between object-based change detection and systematic characterization of complex surface activities, we introduce GeoMorphograms and DeepTopoClustering (DTC), which integrate object-based 4D change analysis with representation learning and hierarchical deep clustering. Experiments on two sandy beach 4D datasets, as well as the combined BeachOBC dataset, demonstrate that DTC can derive a data-driven and expert-supported surface process taxonomy. Quantitative evaluation based on expert annotations exhibits strong agreement between the learned clusters and expert judgments while identifying eight primary surface process types at the highest expert agreement level, achieving an F1 score of 0.78 and an accuracy of 0.92. From the interpretation, the learned hierarchy captures both major erosion-deposition distinctions and finer process variations in duration, magnitude, coherence, compactness, and temporal evolution that are difficult to define manually. Temporal occurrence analysis further shows that the derived process types exhibit distinct dynamics across sites and time periods. Overall, these results demonstrate the potential of hierarchical deep clustering for transforming dense 4D point cloud data into interpretable surface process knowledge. The proposed DTC framework offers a scalable and data-driven way to organize complex spatiotemporal surface activities into a hierarchical taxonomy while allowing the level of detail to be adapted to specific interpretation and application requirements. Future work should extend the framework to a wider range of geomorphic environments, incorporate richer spatial representations, and integrate semantic knowledge through advanced learning strategies. More broadly, this research represents a step toward an automated system for deriving and organizing knowledge from surface activities observed in 4D point clouds, thereby providing higher-level information for understanding underlying environmental processes.

\section*{Acknowledgments}
This research was partly funded by the Deutsche Forschungsgemeinschaft (DFG, German Research Foundation) under project number 535733258 (Extract4D), and partly funded by the Bavarian State Ministry of Science and the Arts in the framework of the bidt Graduate Center for Postdocs (AI4ENV). Daan Hulskemper has been supported by the Nederlandse Organisatie voor Wetenschappelijk Onderzoek (grant no. 20014). The authors gratefully acknowledge Prabin Gyawali, Xiaoyu Huang, and Pim Maydhisudhiwongs for their valuable contributions to the expert annotation process, which was essential for the evaluation of the proposed framework.

\section*{Declaration of generative AI and AI-assisted technologies in the manuscript preparation process}

During the preparation of this manuscript, the author(s) used Grammarly (TUM campus license) and DeepL (free version) to assist with grammar correction and language refinement, and Microsoft Copilot (TUM campus license; GPT-based) to improve phrasing and readability. The author(s) reviewed and edited the output as needed and take full responsibility for the content of the published article. 

\section*{Appendix}\label{sec:appendix}

\begin{figure}[!htb]
\begin{center}
    \includegraphics[width=1.0\columnwidth]{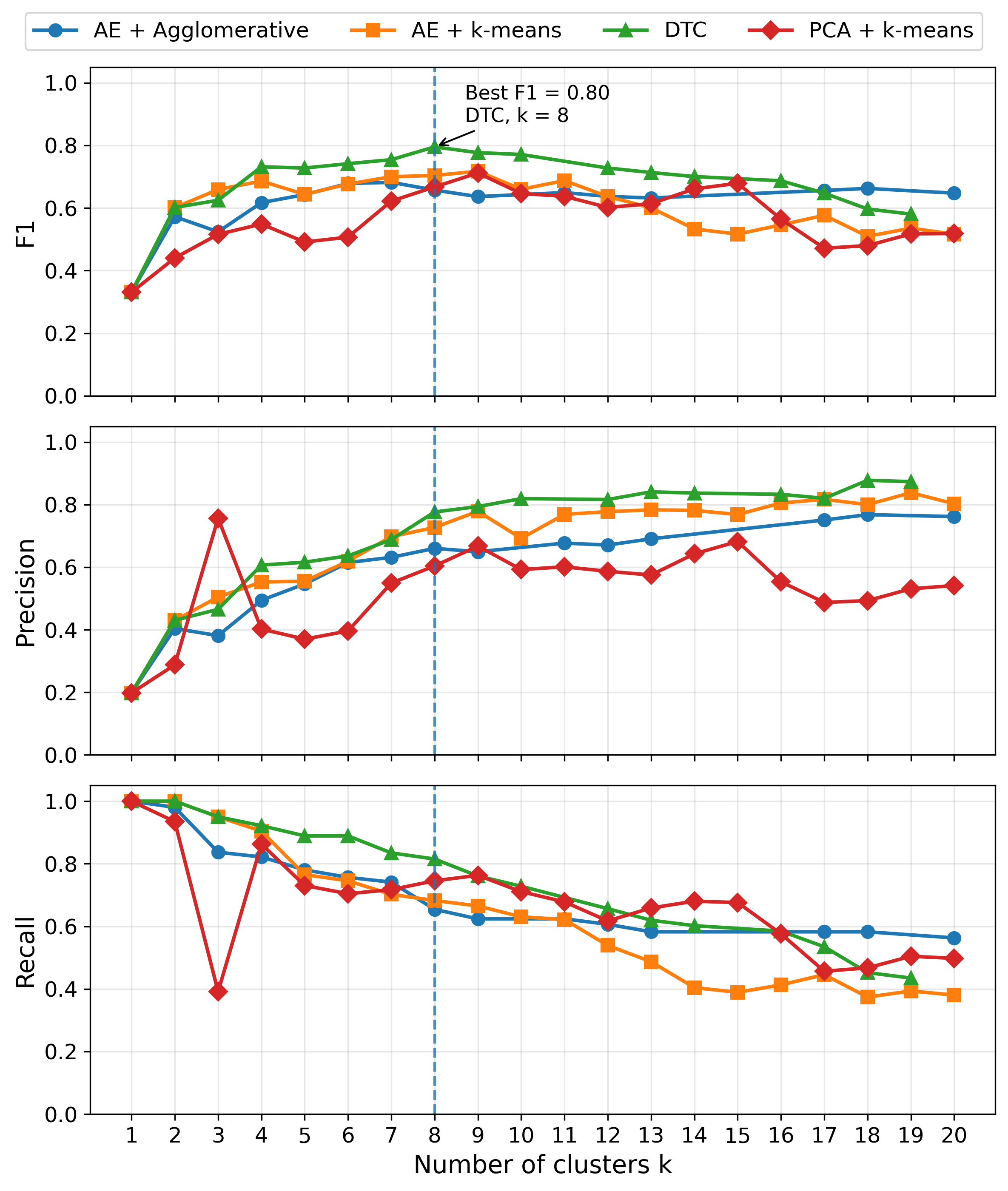}
	\caption{Evaluation metrics computed on pairwise annotations across different numbers of clusters $k$ on the validation set. The dashed line marks the selected level, $k=8$, where the highest overall F1 score is achieved by our approach.}
\label{fig:k-selection-valid}
\end{center}
\end{figure}

\begin{figure*}[!htb]
\begin{center}
    \includegraphics[width=1.0\textwidth]{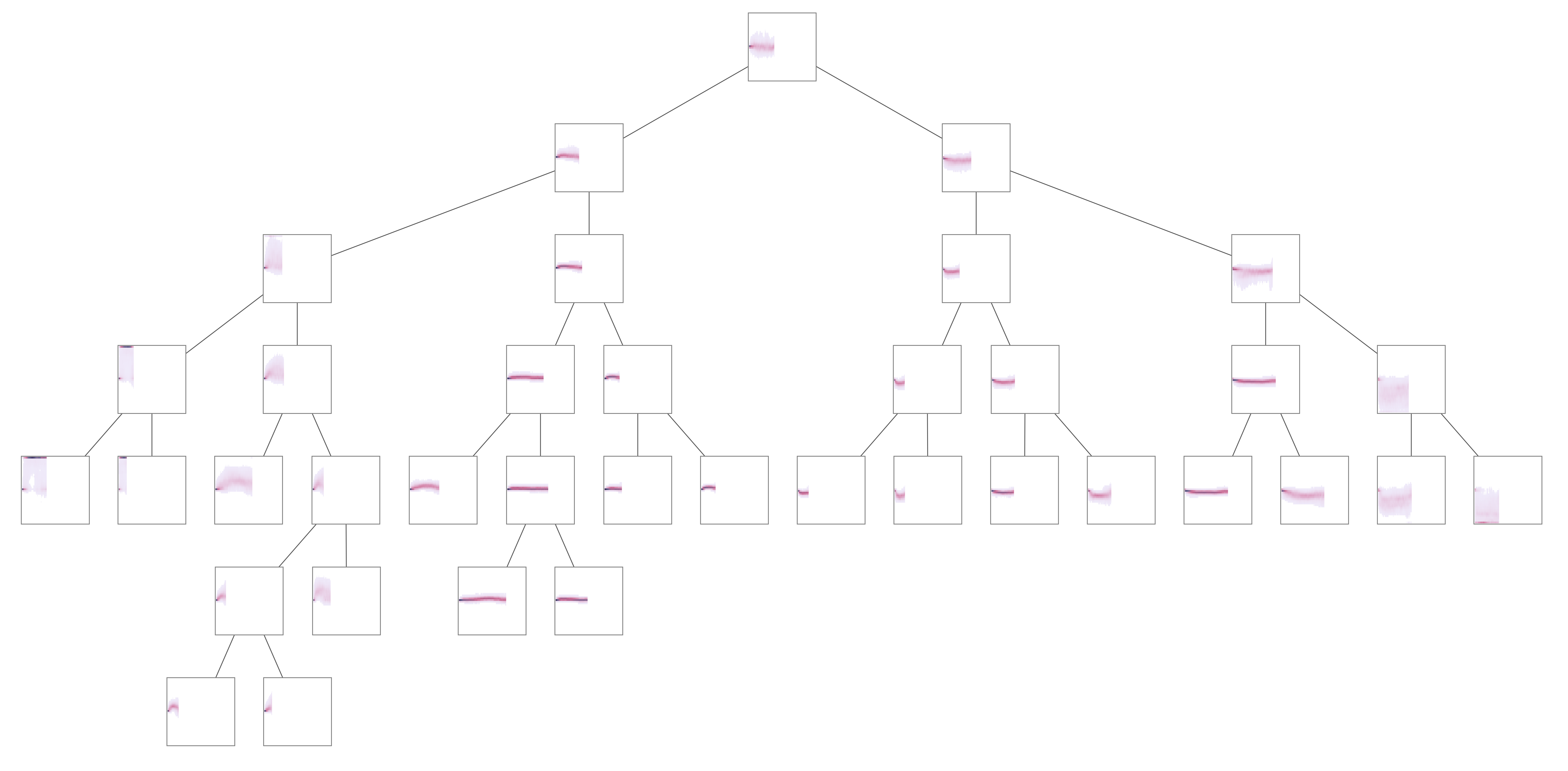}
	\caption{Learned hierarchical tree from the Kijkduin dataset. Each node image represents the decoded GeoMorphogram prototype from the corresponding cluster center, and each edge indicates the parent--child relation between nodes.}
\label{fig:prototype-kijkduin}
\end{center}
\end{figure*}

\section*{Data and code availability}

The permanent laser scanning data underlying this research is published at PANGAEA (\url{https://doi.org/10.1594/PANGAEA.934058}) and 4TU.ResearchData (\url{https://doi.org/10.4121/1aac46fb-7900-4d4c-a099-d2ce354811d2}). The derived 4D-OBCs are available at 4TU.ResearchData (\url{https://doi.org/10.4121/7cf573ec-ce73-4dcf-86ea-553f159c417c}). The code will be published on GitHub after paper acceptance.

\newpage

% \end{thebibliography}
{
    \small
    \bibliographystyle{elsarticle-harv} 
    \bibliography{2026_DeepTopoClustering}
}

\end{document}